# Exploiting Local Optimality in Metaheuristic Search

Fred Glover
Meta-Analytics, Inc.
2605 Stanford Ave
Boulder, CO 80305

fredwglover@yahoo.com

October 2020

**Abstract**

A variety of strategies have been proposed for overcoming local optimality in metaheuristic search. This paper examines characteristics of moves that can be exploited to make good decisions about steps that lead away from a local optimum and then lead toward a new local optimum. We introduce strategies to identify and take advantage of useful features of solution history with an adaptive memory metaheuristic, to provide rules for selecting moves that offer promise for discovering improved local optima.

Our approach uses a new type of adaptive memory based on a construction called exponential extrapolation. The memory operates by means of threshold inequalities that ensure selected moves will not lead to a specified number of most recently encountered local optima. Associated thresholds are embodied in choice rule strategies that further exploit the exponential extrapolation concept. Together these produce a threshold based Alternating Ascent (AA) algorithm that opens a variety of research possibilities for exploration.

*Overview*

Motivated by an early study of metaheuristic search, the Alternating Ascent (AA) Algorithm incorporates new strategies to exploit local optimality within the context of binary combinatorial optimization. The AA Algorithm alternates between an Ascent Phase and a Post-Ascent Phase using thresholds to identify variables to change their values and to transition from one phase to another.

The thresholds embody a form of adaptive memory based on a function called exponential extrapolation, which makes it possible to track the number of times that variables receive their current values in any selected number of most recent local optima. An exponential extrapolation measure EE(j) is associated with a variable $x_j$ that gives rise to a *recency threshold* of the form EE(j) ≥ Threshold(r), which assures that changing the current value for $x_j$ will not duplicate its value in the r most recent local optima. By reference to a standard evaluation Eval(j) for $x_j$ that identifies the change in the objective function when $x_j$ changes its value, and taking advantage of a rudimentary tabu search restriction and aspiration criterion, this in turn gives rise to two status conditions denoted by S1 and S2, where an S1 status identifies a variable that should change the value it received in the most recent local optimum and an S2 status identifies a variable that should retain its value that differs from its value received in the most recent local optimum.

These conditions are additionally exploited using counters StatusCount1 and StatusCount2 of the number of variables that have an S1 and S2 status, embodied in a *trigger threshold* of the form StatusCount1 + StatusCount2 ≥ Trigger. The trigger threshold determines when a new Ascent Phase should be launched by removing all tabu restrictions except the one that caused the threshold to be satisfied. The resulting ascent first reaches a conditional local optimum where the last tabu restriction remains in force, and where it is assured that the solution cannot duplicate any of the r most recent local optima. Then this last restriction is also removed to complete the ascent to a true local optimum, and to begin a new Post-Ascent Phase.

The information underlying these algorithmic components also provides choice rules for selecting variables to change their values during both Ascent and Post-Ascent Phases. Together with the threshold conditions, these rules produce two algorithmic variants, a Single-Pass AA Algorithm and a Double-Pass AA Algorithm. The Single-Pass Algorithm applies the choice rules sequentially while the Double-Pass Algorithm generates information during a First Pass that is used to create new thresholds on the Second Pass to yield more refined choices. Detailed pseudocodes are given for both algorithmic variants and for a special refinement called *Myopic Correction* that modifies previous choices to compensate for limitations in available information.

Numerical examples are given to illustrate the use of exponential extrapolation and the key processes involved in exploiting local optimality via the recency and trigger thresholds.

## 1. Background

An early experiment with metaheuristic search [5] for a class of sequencing problems disclosed that improving moves were more likely to select attributes of optimal solutions than non-improving moves. This was notably reflected in the fact that moves made when approaching a local optimum were more likely to create solutions that shared elements in common with optimal solutions than moves made when retreating from a local optimum. The analysis was conducted using the target analysis approach in which previously discovered (or created) high quality solutions were used to evaluate moves made by alternative decision rules. The study and its findings are amplified in [2], which also provides details of the method used to evaluate and improve the decision rules.

We are motivated by this study to change the rules customarily used by metaheuristic procedures to provide new approaches for responding to local optimality. Our focus is on using adaptive memory strategies that incorporate special threshold inequalities to guide the search.

As a starting point, consider a method that begins from a local optimum and employs rules of the following types, which are commonly employed in a rudimentary form of an adaptive memory tabu search approach.

**Rule 1**: When reaching a local optimum and selecting moves that lead away from this optimum, employ restrictions that temporarily do not allow moves to be reversed and hence that would potentially return to the local optimum.

**Rule 2**: Identify conditions for modifying Rule 1 to permit certain moves to be made that violate the restrictions and allow previous moves to be reversed.

We observe that a simple form of tabu search is often based on a version of these two rules that has two features. We describe these for the purpose of identifying a different way to apply Rules 1 and 2.

**Feature A**: A tenure value is used to prevent a move from being reversed for Tenure iterations, thereby making the reverse move tabu, by reference to the current iteration Iter, by setting

$$\text{TabuIter(ReverseMove)} = \text{Iter} + \text{Tenure}$$

The Reverse Move is then prevented from being made as long as the current value of Iter does not increase beyond an additional Tenure iterations; i.e., as long as Iter satisfies

$$\text{Iter} \leq \text{TabuIter(ReverseMove)}.$$

The tabu restriction on making the Reverse Move therefore expires when Iter > TabuIter(ReverseMove).

The value Tenure can be a fixed value or, in the more usual case, can be a value than varies randomly between chosen values LowTenure and HighTenure. As Iter grows, the "residual tabu tenure" given by TabuIter(ReverseMove) – Iter gradually diminishes until it becomes negative and the reverse move is free to be chosen again.

In the case of a fixed Tenure, the residual tenures associated with moves made more recently will remain in effect longer than those for moves made longer ago, and this relationship will hold approximately for the case of random tenures unless LowTenure and HighTenure are significantly different.

In the case of binary optimization, the reverse of $x_j = 1$ is $x_j = 0$, and vice versa, and hence the form of the reverse move is clear. Then TabuIter(ReverseMove) can be represented simply by TabuIter(j), with the interpretation that Iter ≤ TabuIter(j) means $x_j$ is tabu to change its current value $x_j = x_j^{\#}$ to the reverse (complementary) value $x_j = 1 - x_j^{\#}$.

In other cases, the association between a move and a reverse move is more intricate. For example, in the scheduling study of [5], a move consisted of either an insert move or a swap move, and the reverse move for both cases was defined to consist of moving the first item of the insert or swap again within Tenure iterations. Other definitions of a reverse move in such situations are evidently possible. This leads us to consider the second feature of interest.

**Feature B**: The rule for choosing moves selects a highest evaluation move that is not tabu or that satisfies an aspiration criterion. Since all moves at local optimality cause the objective function to deteriorate or remain unchanged, a highest evaluation move is one that causes the objective function to deteriorate the least. The aspiration criterion most commonly employed considers a Reverse Move to be admissible to be chosen if it leads to a solution better than the best one found so far.

The target analysis experiment described in [5] shows that something about the combination of Feature A and Feature B when moving away from a local optimum tends to produce moves whose attributes do not correspond to those of an optimal solution. Under the assumption that this finding is applicable to other settings, this motivates an examination of versions of Rules 1 and 2 that modify Features A and B to produce a different behavior. In the following sections we focus on the case of binary optimization, as in references [6] to [12].

2. **Alternative forms of Rules 1 and 2**

As a starting point for analyzing conditions that hold at local optimality, we will refer to the use of an evaluation function Eval(j) for a binary variable $x_j$, $j \in N = \{1, \ldots, n\}$, to identify the change in the objective function $x_o$ produced by reversing the assignment $x_j = x_j^{\#}$, where $x_j^{\#}$ is the current value for $x_j$. We suppose the objective is to maximize $x_o$, and hence Eval(j) > 0 corresponds to a move that improves $x_o$. Then all move evaluations are nonpositive when a local optimum is reached, i.e., Eval(j) ≤ 0 for all variables $x_j$.

In the situation where Eval(j) ≤ 0, suppose we assign a tabu tenure to a move that changes $x_j = x_j^\#$ to $x_j = 1 - x_j^\#$ as is customarily done to prevent the move from being immediately reversed. Consider the process that takes place at this point, as the search begins moving away from a local optimum. To begin, all moves selected will consist of reversing values received by variables $x_j = x_j^\#$ in the local optimum to produce new assignments $x_j = 1 - x_j^\#$, and given Eval(j) ≤ 0, these will cause $x_o$ to decrease or remain unchanged.

After reversing an assignment for Eval(j) ≤ 0, the new evaluation Eval(j) will be the negative of its previous value, and hence if Eval(j) began negative it will now be positive. (This is to be expected, since if a move worsened the value of $x_o$, then reversing it will restore the previous value of $x_o$ and therefore be an improving move.) However, the improving move that returns $x_j$ to its previous value will be prevented because of the tabu tenure assigned to it. (The utility of tabu tenures in this respect is to prevent a series of moves that would return to the previous local optimum.) We will build on these simple observations to uncover aspects of adaptive memory choices that have previously been overlooked.

**Overriding tabu restrictions**

As previously intimated, a key question to be addressed in developing an effective algorithm is how to usefully override the customary tabu restriction by freeing certain variables so they are no longer tabu. Accompanying this question is the associated question of identifying the circumstances under which this override should be done. An answer to these questions is suggested by considering the situation where the evaluation for a variable $x_j$ changes from Eval(j) ≤ 0 to Eval(j) > 0 when moving away from a local optimum, without having assigned a new value to $x_j$. We are prompted to ask whether there something noteworthy about this change from a non-improving evaluation to an improving evaluation during a sequence of iterations after reaching a local optimum.

If the current value $x_j^\#$ of $x_j$ is also the value $x_j$ received at the local optimum (when Eval(j) ≤ 0), and if now Eval(j) > 0, then this has the significant feature that the profitable (i.e., improving) move $x_j = 1 - x_j^\#$ gives $x_j$ a different value than it had at the local optimum. If $x_j$ is selected as the variable $x_k$ that changes its value on the current iteration, then by making $x_k$ tabu to change its value, the search cannot return to the local optimum while $x_k$ remains tabu. Consequently, we are motivated to consider the result of freeing the tabu restrictions on all variables $x_j$ except for $x_k$, to launch an ascent in which the procedure cannot return to the previous local optimum. We call the iterations that occur upon launching such an ascent until reaching a new local optimum an Ascent Phase.

Once no more improving moves remain (for the non-tabu variables) in an Ascent Phase, the resulting ascent reaches a *conditional local optimum* (subject to keeping $x_k$ at its new value). At this point, we may remove the tabu restriction on $x_k$ as well, to continue to a solution that is a true local optimum which ends the Ascent Phase. Given that the conditional local optimum does not duplicate the previous local optimum, and that the choice of moves leading to this conditional local optimum is influenced by the value assigned to $x_k$, there is a strong likelihood that the new local optimum will also differ from the previous local optimum.

To exploit this observation, we are presented with the decision of whether to immediately use the change from $Eval(j) \leq 0$ to $Eval(j) > 0$ to trigger an ascent to a conditional local optimum, or whether to wait until more than one variable $x_j$ selected to be $x_k$ has undergone this change before launching such an ascent. We examine this issue in a broader context in the next section.

## 3. A more general procedure for launching a new ascent.

Instead of only considering the most recent local optimum, the issue of identifying a variable $x_j$ to change its value in this local optimum can be generalized to refer to some number Q of the most recent local optima. We describe a way of doing this that makes it possible to maintain appropriate updated information without recording the local optima. This approach, called exponential extrapolation, provides a significant saving of both memory and computation over consulting the actual values of variables in previous local optima.

### 3.1 Exponential extrapolation

The term "exponential extrapolation" is motivated by the term "exponential smoothing," which refers to a procedure that choses a value $\lambda$ between 0 and 1 and uses the simple formula

$$y(q+1) = \lambda y(q) + (1 - \lambda)y(q - 1)$$

to determine the new value of $y(q+1)$ based on the two preceding values $y(q)$ and $y(q-1)$. The procedure can start from chosen values $y(q)$ for $q = 0$ and 1. (More precisely, $y(q+1)$ and $y(q-1)$ refer to forecast values and $y(q)$ refers to an observed value. We do not require this distinction here.)

Exponential extrapolation instead uses the formula, expressed in terms of the weights $w(q)$

$$w(q+1) = \alpha w(q) + \beta q + \gamma \quad (1)$$

where we choose $w(1) = 1$. For simplicity, the parameters $\alpha$, $\beta$ and $\gamma$ may be restricted to $\alpha$ between 1 and 3, and $\beta$ and $\gamma$ between 0 and 3. Even simpler, we will chiefly focus on the special case $\alpha = 2$ and $\beta = \gamma = 0$. The implications for values of $\alpha$ different from 2 are examined in Section 5 and the general case of assigning non-zero values to β and γ is discussed under Variations and Extensions in Appendix 1.

It is possible to establish a connection between exponential extrapolation and exponential smoothing whereby (1) can be seen as a generalization of exponential smoothing, but we will not pursue this here. Exponential smoothing has been applied with tabu search for solving fixed charge network problems in [1], using a different type of design than we use for exploiting exponential extrapolation, but we note that exponential extrapolation affords an alternative to exponential smoothing in the fixed charge setting too.

For the special case of (1) where α > 0 and β = γ = 0 we are particularly interested in the situation where α = 2, to give

$$w(q+1) = 2w(q) \tag{2}$$

This can also be expressed for $q \geq 0$ as $w(q+1) = 2^q$, giving $w(1) = 1$, $w(2) = 2$, $w(3) = 4$, etc., in the familiar expansion of the powers of 2. (In general, the formula $w(q+1) = \alpha w(q)$ can be expressed as $w(q+1) = \alpha^q$ for $q \geq 0$.)

Denote the Q most recent local optima by $x(q)$, $q = 1, \ldots, Q$, writing $x(q) = (x_1^q, x_2^q, \ldots, x_n^q)$. (This notation is more convenient for subsequent purposes than writing $x(q) = (x(1,q), x(2,q), \ldots, x(n,q))$.)

In the following we refer to exponential extrapolation by the acronym EE and are interested in two weighted EE values for each variable $x_j$: EE1(j) which weights the values $x_j^q$ that equal 1 in the vectors $x(q)$, $q = 1$ to $Q$, and EE0(j) which weights the values $x_j^q$ that equal 0 in these vectors.

Then for each $x_j$ we define

$$EE1(j) = \sum(w(q)x_j^q : q = 1, \ldots, Q) \quad (= \sum(2^{q-1} x_j^q : q = 1, \ldots, Q)) \tag{3.1}$$

and

$$EE0(j) = \sum(w(q)(1 - x_j^q) : q = 1, \ldots, Q) \tag{3.2}$$

A useful shortcut for equation (3.2) is given by defining

$$EEbase = \sum(w(q) : q = 1, \ldots, Q) \quad (= \sum(2^{q-1} : q = 1, \ldots, Q))$$

Then EE0(j) can be written

$$EE0(j) = EEbase - EE1(j) \tag{3.3}$$

By well-known properties of binary sequences, the parameter $\alpha = 2$ that gives (2) implies EEbase $= 2^Q - 1$. Moreover, for each $q = 2$ to $Q$, $w(q) = \sum(w(h): h = 1, \ldots, q - 1) + 1 = 2^{q-1}$. Hence $w(q)$ is greater than the sum of all weights $w(h)$ for $h < q$, and as a special case $w(Q) > \sum(w(q): q = 1, \ldots, Q - 1)$. (To illustrate, for $Q = 4$, we have EEbase $= 1 + 2 + 4 + 8 = 15 = 2^4 - 1$ and $w(4) = 2^3 = 8$.)

Consequently, the value EE1(j) will be larger when $x_j^Q = 1$ than it will be when $x_j^Q = 0$, regardless of the values $w(q)$ for $q < Q$. Another way of expressing this is that (2) creates a lexicographic ordering of the binary value assignments to the variables $x_j^q$, where the value EE1(j) $= \sum w(q)x_j^q: q = 1, \ldots, Q)$ is larger as the vector $V(j) = (x_j^Q, x_j^{Q-1}, \ldots, x_j^1)$ increases lexicographically.

These formalisms can be illustrated by the following example. Suppose $Q = 6$ and the vector $V(j)$ of the most recent values $x_j = x_j^q$, for $q = 1$ to 6 is $(1, 1, 1, 1, 1, 1)$, hence all 6 of these most recent $x_j$ values $= 1$. The corresponding weighted values $w(q)x_j^q$ (for $w(q) = 2^{q-1}$), written in reverse order so the most recent value for $Q = 6$ appears first, are $(32, 16, 8, 4, 2, 1)$. The EE value EE1(j) $= \sum w(q)x_j^q: q = 1, \ldots, Q)$, likewise written in reverse order, is $32 + 16 + \ldots + 1 = 63$ ($= 2^{Q-1} = 2^6 - 1$) (which is the maximum possible value for EE1(j) due to the fact that $x_j = 1$ in all of these Q solutions).

But if $x_j = 1$ only for the most recent solution (i.e., $x_j^Q = 1$ for $Q = 6$), and all other $x_j^q$ values for $q < 6$ are 0, which gives $V(j) = (1, 0, 0, 0, 0, 0)$ when written in the order $(x_j^Q, x_j^{Q-1}, \ldots, x_j^1)$, the resulting EE1(j) value $= 32$ is greater than if $x_j^Q = 0$ but $x_j^q = 1$ for all $q < 6$, giving $(0, 1, 1, 1, 1, 1)$ with EE1(j) $= 31$ ($= 0 + 16 + 8 + 4 + \ldots + 1 = 2^6 - 1$). Similarly, $V(j) = (1, 0, 1, 0, 0, 0)$ gives EE1(j) $= 40$, which is greater than EE1(j) $= 39$ for the vector $V(j) = (1, 0, 0, 1, 1, 1)$ (a consequence of the fact that the vector $(1, 0, 1, 0, 0, 0)$ is lexicographically larger than the vector $(1, 0, 0, 1, 1, 1)$).

For our purposes, this means that by using the EE value EE1(j), the most recent local optimum recorded ($x(Q)$) will dominate any combination of all other local optima, and the second most recent local optimum ($x(Q - 1)$) will dominate any combination of all local optima preceding it, and so forth. A useful implication is that if we want to make sure that $x_j = 1$ in each of the 3 most recent local optima, which receive the three largest weights $2^5$, $2^4$ and $2^3$ in the preceding example, we would simply require that EE1(j) $\geq 56$ ($= 32 + 16 + 8$).

We denote the *threshold value* that sums the r largest weights by

$$\text{Threshold}(r) = 2^{Q-1} + 2^{Q-2} + \ldots + 2^{Q-r}$$

Let $x_j^{\#}$ denote the current value of $x_j$, and define

$$EE(j) = EE1(j) \text{ if } x_j^{\#} = 1 \text{ and } EE(j) = EE0(j) \text{ if } x_j^{\#} = 0.$$

Then, in general, we can require that we only select a variable $x_j$ to change its value from $x_j^\#$ to $1 - x_j^\#$ if $x_j = x_j^\#$ in the r most recent local optima by stipulating

$$EE(j) \geq \text{Threshold}(r) \qquad (4)$$

We call the inequality (4) the *recency threshold.*

We treat recency threshold as embodying the two inequalities

$$EE1(j) \geq \text{Threshold}(r) \qquad (4.1)$$

$$EE0(j) \geq \text{Threshold}(r) \qquad (4.2)$$

where (4.1) applies to $x_j = 1$ and requires that $x_j$ cannot be chosen to change from 1 to 0 unless $x_j$ also equals 1 in each of the r most recent local optima, and (4.2) applies to $x_j = 0$ and requires that $x_j$ cannot be chosen to change from 0 to 1 unless $x_j$ also equals 0 in each of the r most recent local optima.

In short, by requiring the recency threshold (4) to be satisfied (for any choice of the index j, and for a specified value $x_j = x_j^\#$), we assure that we will not risk duplicating any of the r most recent local optima by changing $x_j$ to equal $1 - x_j^\#$. The utility of this requirement is that we do not need to record the most recent local optima to verify – or compel – that $x_j = 1$ or 0 in any specified number r of these most recent solutions. All that is necessary is to specify that EE1(j) satisfy (4.1) or that EE0(j) satisfy (4.2), according to whether $x_j^\# = 1$ or 0.

We will have occasion to consider the recency threshold for a special case where EE1(j) and EE0(j) are not determined by the value $x_k^\# = 1$ or 0, but are instead determined by the complementary value of $x_j$, i.e., where EE1(j) refers to $x_j = 1 - x_j^\# = 1$ ($x_j^\# = 0$) and EE0(j) refers to $x_j = 1 - x_j^\# = 0$ ($x_j^\# = 1$) . Let $EE^c(j)$ refer to this complementary form of EE(j) determined relative to the complement $x_j = 1 - x_j^\#$. As a result of (3.3), which gives EE0(j) = EEbase – EE1(j), we have

$$EE^c(j) = EEbase - EE(j)$$

and consequently, the recency threshold (4) is equivalent to

$$EE^c(j) \leq EEbase - \text{Threshold}(r) \qquad (4^c)$$

These relationships involving complements are elaborated in Appendix 3.

Now we show how to conveniently calculate and update the value EE(j) by showing how to update EE1(j) of (3.1) (which automatically gives EE0(j) = EEbase – EE1(j) by (3.3) and in turn gives the values of each of the other terms defined by the preceding relationships).

## 3.2 An inductive updating formula

We consider the situation where the number of local optima Q constant. We write EE1(j) with the most recent q value Q first:

$$EE1(j) = w(Q)x_j^Q + w(Q-1)x_j^{Q-1} + \ldots + w(1)x_j^1$$

Holding Q constant means that when a new local optimum is generated to become the new x(Q), the previous EE1(j) value, denoted by EE1(j):p (letting p denote "previous") may be expressed as

$$EE1(j){:}p = w(Q)x_j^Q{:}p + w(Q-1)x_j^{Q-1}{:}p + \ldots + w(1)x_j^1{:}p$$

Then a useful way to update EE1(j) (without having stored the x(q) vectors and their $x_j^q$ values from earlier iterations) is given by

$$EE1(j) = 2^{Q-1}x_j^Q + EE1(j){:}p/2 \qquad (5)$$

(The form of this update for an arbitrary $\alpha > 0$ is given by $EE1(j) = \alpha^{Q-1}x_j^Q + EE1(j){:}p\,/\alpha$.)

To establish this result, we use the fact that $x_j^q{:}p$ would be the value now indexed as $x_j^{q-1}$. (Hence, what would now be called $x_j^0$ was the value $x_j^1{:}p$ from before). So, we can also write

$$EE1(j){:}p = w(Q)x_j^{Q-1} + w(Q-1)x_j^{Q-2} + \ldots + w(2)x_j^1 + w(1)x_j^0$$

Now, however, these previous $x_j$ values should receive new weights, from w(Q – 1) to w(0), where w(0) = 0 (since the old $x_j^1$:p, now denoted $x_j^0$ will not be part of the current EE1(j)). Hence, the updated value EE1(j):u of EE1(j):p will be given by

$$EE1(j){:}u = w(Q-1)x_j^{Q-1} + w(Q-2)x_j^{Q-2} + \ldots + w(1)x_j^1 + w(0)x_j^0$$

By the formula w(q+1) = 2w(q), this gives EE1(j):u = EE1(j):p/2 as stipulated in (5).

If we use integer arithmetic that rounds fractional values less than 1 down to 0, this division by 2 automatically has the same effect as setting w(0) = 0, as desired, since $w(0)x_j^0 = w(1)x_j^0/2$. Given that w(1) = 1 and $x_j^0$ (the previous $x_j^1$ value) is at most 1, we have $w(0)x_j^0 \leq .5$  Integer arithmetic will cause this fractional term to be dropped, and EE1(j):u will only refer to the Q most recent $x_j$ values $x_j^q$, q = 1 to Q. Finally, the new $x_j^Q$ value receives the weight $w(Q) = 2^{Q-1}$ to give the updated form of EE1(j) as specified in (5).

If real (floating point) arithmetic is used instead of integer arithmetic, the declining influence of earlier $x_j$ values will proceed in the same manner as if Q had been chosen to be larger, or equivalently as if we allowed q to become negative, with each weight w(q – 1) being half of the weight w(q), hence yielding w(0) = 1/2, w(– 1) = 1/4, w(– 2) = 1/8, … and so forth. By the relationship w(q) = $2^{q-1}$ this corresponds to the weights $2^{-1}$, $2^{-2}$, $2^{-3}$, …. The values q = 0, – 1, – 2 … need not be created or accessed, of course, since they are merely a notational convention to convey how using real arithmetic will have the same effect as permitting Q to be larger. This can be relevant when using α values different than 2, as discussed in Appendix 2.

The inductive update conveniently permits us to start with Q at its maximum desired value and use the formula (5) at each iteration of generating a new local optimum to update EE1(j). Until Q local optima have been generated, EE1(j) will not refer to terms that go all the way back to q = 1. For example, after generating s local optima for s < Q, EE1(j) determined by (5) will effectively yield EE1(j) = $2^{Q-1}x_j^Q + 2^{Q-2}x_j^{Q-1} + \ldots + 2^{Q-s}x_j^{Q-s+1}$

Hence, if we want to apply the recency threshold to assure $x_j = 1$ in the r most recent local optima, we must remember that r cannot exceed s. Fortunately, the inductive update handles this automatically.

In the situation where s < Q, the EEbase value is given by EEbase = $2^{Q-1} + 2^{Q-2} + \ldots + 2^{Q-s}$. EEbase can be computed inductively as well as EE1(j), to automatically give the correct value for s < Q, by initializing

$$\text{EEbase} = 2^{Q-1} \tag{5.1}$$

and then, each time a new local optimum is obtained, by setting

$$\text{EEbase} = 2^{Q-1} + \text{EEbase}/2 \tag{5.2}$$

After generating Q local optima, then EEbase = ∑w(q): q = 1, …, Q) = $2^{Q-1} + 2^{Q-2} + \ldots + 1 = 2^Q - 1$, as previously observed. This EEbase value applies under the assumption that integer arithmetic is used to compute EE1(j), while EE0(j) is determined from EE1(j) using (3.3). If real (floating point) arithmetic is used for computing EE1(j) by (5) and EEbase by (5.1) and (5.2), then EEbase and EE1(j) will both implicitly to go beyond Q terms to include the weights 1/2, 1/4, 1/8, … as previously noted.

In the same way, the value of Threshold(r) can only refer to the s most recent local optima when s < r. We can inductively update Threshold(r) by letting Threshold$_o$ denote the constant value (once r is selected) given by

$$\text{Threshold}_o = \text{Threshold}(r) = 2^{Q-1} + 2^{Q-2} + \ldots + 2^{Q-r} \tag{5.3}$$

Then the current value, ThresholdR, for Threshold(r) is given by

$$\text{ThresholdR} = \text{Min}(\text{EEbase}, \text{Threshold}_o) \tag{5.4}$$

These relationships provide the foundation for two main strategies for exploiting local optimality.

### 3.3 Exploiting local optimality based on exponential extrapolation

We refer to the iterations that begin upon reaching a local optimum with an Ascent Phase as a Post-Ascent Phase. We are interested in two principal strategies to guide the Post-Ascent Phase. Each depends on the existence of an opportunity to interrupt the search by removing tabu restrictions and then to proceed to a new local optimum.

**Strategy S1.**

We refer to three key conditions that may be satisfied by a variable $x_j$ after a local optimum is reached (where, to begin, $Eval(j) \leq 0$ for all variables $x_j$).

(i) The current value $x_j^\#$ of $x_j$ is the same as the value of $x_j$ in the most recent local optimum $x(Q)$ (i.e., $x_j^\# = x_j^Q$)).
(ii) The evaluation of $x_j$ is currently profitable (i.e., $Eval(j) > 0$) and $x_j$ is not tabu.
(iii) The evaluation of $x_j$ is currently profitable and $x_j$ satisfies the recency threshold $EE(j) \geq Threshold(r)$ (i.e., $EE1(j) \geq Threshold(r)$ if $x_j^\# = 1$ and $EE0(j) \geq Threshold(r)$ if $x_j^\# = 0$). .

A variable $x_j$ that satisfies (i) and (ii) will be said to have an *S+ status*. The "S" in S+ simply refers to "strategy," while the "+"refers to the fact that $Eval(j) > 0$, which implies that changing the value of $x_j$ will produce an improvement in the objective function $x_o$. A variable with an S+ status will be given a higher priority to be selected as the variable $x_k$ (to receive the new value $x_k = 1 - x_k^\#$) than a variable that does not have an S+ status (i.e., a non-tabu variable with $Eval(j) \leq 0$). In other words, any non-tabu variable with a profitable evaluation has a higher priority of receiving a new value than one with a non-profitable evaluation.

A variable $x_j$ that satisfies all three conditions (i), (ii) and (iii) (or equivalently, the conditions (i) and (iii)) will be said to have an S1 status. The S1 status dominates the S+ status by having a higher priority to be chosen as the variable $x_k$ that receives a new value. Note this implies that the recency threshold acts like an aspiration criterion that overrides a tabu restriction to allow a variable $x_j$ to be selected when $Eval(j) > 0$. We do not allow this to happen when $Eval(j) \leq 0$.

The importance of the S1 status is that it means that the choice of $x_j$ to become $x_k$ can participate in a decision to trigger an assent to a new local optimum, as described below. The way that the S1 status contributes to this process is as follows.

The recency threshold $EE(j) \geq Threshold(r)$ ($EE1(j) \geq Threshold(r)$ if $x_j^\# = 1$ and $EE0(j) \geq Threshold(r)$ if $x_j^\# = 0$) implies that $x_j = x_j^\#$ in each of the r most recent local optima. Hence if $x_j$ is selected as $x_k$ to set $x_k = 1 - x_k^\#$ on the current move, the resulting solution cannot move toward any of these local optima. An S1 status is *realized* by assigning $x_k$ its new profitable value, thereby causing its new $x_k^\#$ value to be the complement of its current value.

The value of r must be chosen large enough (analogous to the choice of a tabu tenure in tabu search) to drive the search away from an appropriate number of previous local optima. At the same time, the inequality $EE(j) \geq Threshold(r)$ is stronger than necessary to avoid visiting these r local optima. Consequently, it can be preferable to avoid making r too large, which may unduly restrict the new solutions that can be reached. (This observation also suggests that it may be valuable to explore options for setting α smaller than 2. This issue is examined in Section 4.)

The principal observation to be made at present is that the greater the number of variables $x_j$ that receive an S1 status and that have been chosen to be $x_k$, the greater is the motive for triggering an ascent to a new local optimum.

**Strategy S2.**

The second strategy arises where a variable $x_j$ satisfies the following conditions:

(iv) The current value $x_j^\#$ of $x_j$ differs from the value of $x_j$ in the most recent local optimum $x(Q)$ (i.e., $x_j^\# \neq x_j^Q$).
(v) The current evaluation of $x_j$ is unprofitable ($Eval(j) < 0$).
(vi) The recency threshold $EE1(j) \geq Threshold(r)$ is satisfied if $x_j^Q = 1$ and the recency threshold $EE0(j) \geq Threshold(r)$ is satisfied if $x_j^Q = 0$.

Note that a variable $x_j$ that satisfies $x_j^\# \neq x_j^Q$ will have been made tabu after reaching the local optimum $x(Q)$ by the customary approach of making any variable tabu when it changes its value. The variable will continue to be tabu unless its Tenure value has expired in the interim after receiving this new value. We will suppose that we make Tenure large enough to avoid this eventuality.

When the move occurred to change $x_j$'s value to $x_j^Q$, the original evaluation $Eval(j) \leq 0$ would have reversed its sign to become $Eval(j) \geq 0$ (which, if $> 0$, would have made $x_j$ profitable to change back to its previous value except for the tabu restriction). The current evaluation $Eval(j) < 0$ by (v) contrasts with the situation that created (iv). This current evaluation is therefore consistent with considering the previous change of $x_j$ to have been a profitable move rather than a non-profitable move (since the negation of $Eval(j)$ for a profitable move would cause $Eval(j) < 0$ as in (v)).

A variable $x_j$ that satisfies all three conditions (iv), (v) and (vi) will be said to have an S2 status. The importance of the S2 status is that, like an S1 status, it qualifies $x_j$ participate in a decision to trigger an assent to a new local optimum. (This results from the fact that the move that has given $x_j$ its new value $x_j^\#$ receives an evaluation as if it had originally been profitable and, in addition, causes $x_j$ to satisfy the recency threshold for $x_j = x_j^Q$.) In sum, as we indicate below, once a sufficient number of variables have either an S1 status or an S2 status, then these variables activate an Ascent Phase that proceeds to a new local optimum.

It may be seen that (vi) is identical to (iii) by noting that $x_j^\#$ in (iii) also corresponds to $x_j^Q$. The difference between (iii) and (vi) is that the value $x_j^Q$ in (vi) differs from the current value $x_j^\#$ for $x_j$. Because variables are made tabu when they are assigned a new value after reaching a local optimum, this implies that an S1 status deals with the case where $x_j$ has not yet changed its local optimum value and an S2 status deals with the case where such a change has already occurred.

In short, the new value not yet assigned to $x_j$ in (iii) and the new value already assigned to $x_j$ in (vi) must be different than the value of $x_j$ in each of the r most recent local optima, and hence by receiving this value, the current solution moves in a direction away from these most recent local optima.

It should also be observed that by the convention that EE(j) is determined relative to the current value $x_j^\#$ of $x_j$ instead of the value of $x_j^Q$ as in (vi), which causes EE1(j) to refer to $x_j^\# = 1$ and EE0(j) to refer to $x_j^\# = 0$, the values EE1(j) and EE0(j) in (vi) refer to the complements EE1$^c$(j) and EE0$^c$(j). As shown earlier in (4$^c$), this implies that when EE1(j) refers to $x_j^\# = 1$ and EE0(j) refers to $x_j^\# = 0$, the inequality of (vi) becomes

$$EE(j) \leq EEbase - Threshold(r).$$

This is relevant for choice rules, since it further implies that the value of EE(j) determined relative to $x_j^\#$ will be relatively small when (vi) holds relative to $x_j^Q$, and in general, just as EE(j) benefits from being larger relative to $x_j^\#$ in order for the recency threshold EE1(j) $\geq$ Threshold(r) to hold (which establishes S1 status when Eval(j) > 0 and makes it desirable to select $x_j$ to become $x_k$ and change its value), the corresponding inequality EE(j) $\leq$ EEbase – Threshold(r) from (vi) shows that when Eval(j) $\leq$ 0, a smaller EE(j) is associated with a case where it is undesirable to select $x_j$ to change its value. Consequently, regardless of the sign of Eval(j), there is a motivation to favor a larger EE(j) when choosing a variable $x_j$ to change its value.

### 3.4 Applying Strategies S1 and S2 to trigger an Ascent Phase

We introduce a variable StatusCount that counts the number of variables $x_j$ that have either an S1 or S2 status. As illustrated in the next section, StatusCount may decrease on some iterations, because a variable $x_j$ with an S1 or S2 status may lose this status after a move involving a different variable is made. The following *trigger threshold* provides a rule for launching an ascent to a new local optimum

$$\text{StatusCount} \geq \text{Trigger} \qquad (6)$$

Suggested values for the Trigger parameter are from 5 to 9 (subject to modification by experiment).

Once the trigger threshold is satisfied, we know that if we continue to hold any of the $x_j$ variables with an S1 or S2 status at its current value then we cannot duplicate any of the r most recent local optima. Thus we can select the last of these variables to remain tabu and, as intimated earlier, remove the tabu restrictions on all other variables and freely choose those with positive evaluations to ascend to a *conditional local optimum* – a solution that is locally optimal subject to retaining the tabu restriction on the last variable. Upon reaching the conditional local optimum, the tabu restriction on the remaining variable is likewise removed and the method proceeds to a true local optimum. (A natural variation is to allow all tabu restrictions to be removed from the beginning of the ascent in the expectation that the Trigger threshold will create a high probability that the new local optimum reached will not duplicate any of the r most recent local optima. A contrasting variation would retain all variables with an S1 and S2 status tabu and release them all together from their tabu restrictions at the conditional local optimum.)

Once more, we emphasize that it is not necessary to store the local optima, and this includes the most recent local optimum x(Q) in spite of referring to its value in (iv) of Strategy S2. Under the default assumption that EE(j) is based on $x_j = x_j^\#$, x(Q) may be "forgotten" after updating EE1(j) when the solution x(Q) is obtained, because we know the current value of $x_j$ yields $x_j^\# = x_j^Q$ when EE(j) $\geq 2^{Q-1}$ and yields $x_j^\# \neq x_j^Q$ when EE(j) $< 2^{Q-1}$. (This follows from the relationships identified in Section 3.1.) In fact, when $x_j^\# \neq x_j^Q$, EE(j) will be typically be much smaller than $2^{Q-1}$ by the observations in Appendix 3.

The next section provides an extended example of how these relationships are exploited.

## 4. Extended Illustration of Exploiting Strategies S1 and S2.

We illustrate how the preceding data structures can be used to implement the two strategies S1 and S2 by an example of the steps following an Ascent Phase to move away from the most recent local optimum in a Post-Ascent Phase. We choose Q = 4 to identify the most recent local optimum x(Q). For clarity, the following Working Table shows all 4 of the most recent local optima x(1) to x(4), although it isn't necessary to keep a record of these solutions in order to execute the method. The Working Table also shows the weighted sums EE1(j) and EE0(j) whose values appear just beneath the solution x(Q) = x(4) (the last solution shown).

As a prelude to discussing the moves shown in the Working Table, recall that Strategies S1 and S2 always use (a) EE(j) = EE1(j) when the most recent solution x(Q) has $x_j^Q = 1$ and (b) EE(j) = EE0(j) when the most recent solution x(Q) has $x_j^Q = 0$. We emphasize this correspondence in the Working Table with a yellow highlight for case (a) and a green highlight for case (b). The values shown in the row for "EE1(j) | EE0(j)" in the Working Table therefore refer to EE1(j) when $x_j^Q =$

1 and refer to EE0(j) when $x_j^Q = 0$ in the local optimum x(4). (This correspondence can be confirmed by computing EE1(j) and EE0(j) using (3.1) and (3.3).)

We have chosen a Trigger value of 3 for the trigger threshold StatusCount ≥ Trigger in (6) that launches an ascent to a new local optimum. Details of the following Working Table are discussed immediately after the table.

**Working Table**

| q | w(q) | x(q) | $x_1$ | $x_2$ | $x_3$ | $x_4$ | $x_5$ | $x_6$ | $x_7$ | $x_8$ | $x_9$ | $x_{10}$ | |
|---|---|---|---|---|---|---|---|---|---|---|---|---|---|
| 1 | 1 | x(1) | 1 | 1 | 1 | 1 | 0 | 0 | 0 | 1 | 1 | 1 | |
| 2 | 2 | x(2) | 0 | 1 | 0 | 1 | 1 | 0 | 1 | 1 | 0 | 1 | |
| 3 | 4 | x(3) | 0 | 1 | 0 | 0 | 0 | 0 | 1 | 1 | 1 | 0 | |
| 4 | 8 | x(4) | 1 | 1 | 0 | 0 | 0 | 0 | $1^T$ | 1 | $1^T$ | 0 | |
| EE1(j) \| EE0(j) | | | 9 | 15* | 14* | 12 | 13 | 15* | 14* | 15* | 13 | 13 | * for ≥ 14 |
| Move | | | (Begin a Post-Ascent Phase) | | | | | | | | | | StatusCount |
| 1 | | | 0 | | | | | | | | | | |
| 2 | | | | 0 | | | | | | | | | |
| 3 | | | | | 1 | | | | | | | | |
| 4 | | | | S2 | | 1 | | | | | | | 1 |
| 5 | | | | | | | 1 | | S1 | | S+ | | 2 |
| 6 | | | | X | | | | | 0 | | X | | 1 |
| 7 | | | | S2 | | | | 1 | | | S+ | | 2 |
| 8 | | | | | S2 | | | | | | 1 | | 3 |
| | | | Launch a new Ascent Phase | | | | | | | | | | |
| q | w(q) | x(q) | (Obtain a new local optimum x(4))[1] | | | | | | | | | | |
| 4 | 8 | x(4) | 1 | 0 | 1 | 0 | 0 | 1 | 1 | 1 | 1 | 1 | |
| EE1(j) \| EE0(j) | | | 12 | 8 | 8 | 14* | 14* | 8 | 15* | 15* | 14* | 8 | (* for ≥ 14) |

[1]The new x(1), x(2), x(3) are the previous solutions x(2), x(3), x(4).

**Working Table Explanation**

The method begins the Post-Ascent Phase with the most recent local optimum $x(4) = x(Q)$, whose values $x_j = 1$ are highlighted in yellow and whose values $x_j = 0$ are highlighted in green. The corresponding values for $EE(j)$ ($EE1(j)$ and $EE0(j)$) are likewise respectively highlighted in yellow and green.

An asterisk (*) has been attached to each value $EE(j)$ value that satisfies $EE(j) \geq Threshold(r)$, which is relevant for identifying moves with an S1 or S2 status that will cause StatusCount to change. Here we have chosen $r = 3$, yielding $Threshold(3) = 8 + 4 + 2 = 14$ (the sum of the three largest $w(q)$ values). Hence an asterisk is attached to each $EE(j)$ value that is at least 14.

For this example, we do not bother to specify the choice rule used to select variables $x_j$ to set equal to 0 or 1. A discussion of choice rules is given in Section 6. As a basis for tracking the choices made, recall that $Eval(j)$ is nonpositive for all variables at a local optimum. Hence the first choice of a variable $x_j$ to change its value after reaching the local optimum $x(4)$ will be for a variable with $Eval(j) \leq 0$. Each choice of such a variable will reverse the sign of $Eval(j)$ to produce $Eval(j) \geq 0$, as noted in condition (iv) of the S2 strategy.

As previously noted, we assume that each variable selected to change its value is made tabu to prevent a move that changes the variable back to its previous value. We also assume in the present example that the variables $x_7$ and $x_9$ are tabu in the local optimum $x(4)$, as indicated by the superscript T attached to the values for these variables in the $x(4)$ row. (Variables may receive a tabu restriction in this way by a rule that, upon obtaining a local optimum, selects some number of the variables that were most recently assigned values leading to this local optimum to be tabu. Here we may suppose $x_7$ and $x_9$ were the last two variables to be assigned their current values to reach this local optimum. It would also be possible to apply a rule that does not make any variables in the local optimum tabu. However, we include the situation with $x_7$ and $x_9$ tabu to increase the scope of the illustration.)

After one or more moves have been made, the evaluation for one of the previously selected variables $x_j$ can change. This can be the basis for identifying a $x_j$ that qualifies to receive an S2 status because its evaluation has changed to become $Eval(j) \leq 0$. Similarly, the evaluation of a variable $x_j$ that has not previously been selected can change from $Eval(j) \leq 0$ to $Eval(j) > 0$, qualifying $x_j$ to receive an S+ or an S1 status.

*Description of Successive Moves*

As shown in the Working Table, Move 1 selects $x_1$ as the first variable to become $x_k$ to change its value, changing $x_1 = 1$ to $x_1 = 0$, with its new value 0 shown in the row for Move 1.

Similarly, Move 2 chooses $x_2$ to change from 1 to 0, as indicated by the value 0 shown in the row for Move 2, and Move 3 chooses $x_3$ to change from 0 to 1, as indicated by the value 1 shown in the row for Move 3.

The next move, Move 4, chooses $x_4$ to change its current value, yielding $x_4 = 1$, indicated by the value 1 shown in the row for Move 4. In addition, the symbol S2 for Strategy S2 is inserted in this row in the column for $x_2$ to disclose that the result of setting $x_4 = 1$ has changed a current evaluation Eval(2) > 0 for $x_2$ to a new evaluation Eval(2) < 0, and in addition EE1(2) ≥ 14, as indicated by asterisk attached to the value 15 for EE1(2) (where EE(2) = EE1(2) is defined in relation to $x_2 = x_2^Q = 1$), shows that $x_2$ qualifies for the S2 status of Strategy S2. As a result, the StatusCount in the far-right column of the table is incremented to 1 (from an implicit initial value of 0).

Move 5 selects $x_5$ to change from 0 to 1, and now this move causes $x_7$ and $x_{10}$, which have not yet changed their values in the local optimum x(4), to receive new evaluations Eval(7) and Eval(10) > 0, hence making them profitable and qualifying them for the S+ status. In addition, $x_7$ satisfies the recency threshold and hence qualifies for the S1 status. Thus, we show S1 in the column for $x_7$ and S+ in the column for $x_{10}$, and the S1 status for $x_7$ results in incrementing StatusCount to 2 in the far-right column.

The status S1 and S+ for $x_7$ and $x_{10}$ (which also identifies them as improving moves) give both variables priority to be selected to change their values. Since the S1 status is higher than the S+ status, we select $x_7$ to change its value from 1 to 0 in Move 6. We make this move in spite of the fact that $x_7$ begins tabu (as indicated by the superscript T attached to its value in x(4)), because the S1 priority also overrules the tabu status.

Move 6 to set $x_7 = 0$ additionally has two other consequences in this example. The X's in the columns for $x_2$ and $x_{10}$ are used to indicate that S2 and S+ statuses of these variables have been cancelled because of the move setting $x_7 = 0$ – a situation indicating that setting $x_7 = 0$ causes Eval(2) to become positive and Eval(10) to become nonpositive. Because of cancelling the S2 status of $x_2$, StatusCount is reduced from 2 to 1.

There now remain three variables that are not tabu, $x_6$, $x_8$ and $x_{10}$ (unless the tabu tenure attached to $x_9$ is small enough that the tabu status of $x_9$ has expired). Move 7 selects $x_6$ to change its value from 0 to 1. According to the table, this move causes Eval(2) and Eval(10) once again to become nonpositive and positive, respectively, and consequently reinstates their S2 and S+ status that was cancelled on the previous move. (Such a rapid fluctuation of the nonpositive and positive evaluaions of variables may be unlikely, but we show such a change to illustrate conditions that potentially may happen.) The recovery of the S2 status by $x_2$ causes StatusCount again to grow to 2.

Variable $x_{10}$ with its S+ status now has priority above other variables to be chosen as $x_k$, and the assignment $x_{10} = 1$ occurs in Move 8. This move causes $x_3$ to receive an S2 status (by changing

Eval(3) > 0 back to Eval(3) ≤ 0 and observing EE1(2) = 15 which is larger enough for $x_2$ to satisfy the recency threshold). Now StatusCount increases again, to equal 3.

Since we have chosen Trigger to be 3, the trigger threshold StatusCount ≥ Trigger is now satisfied and the ascent to a new local optimum is launched. The variable $x_3$ is held tabu until reaching a conditional local optimum, and then its tabu restriction is released as well to proceed to a true local optimum.

The next to last row of the Working Table identifies the new local optimum, again designated x(4) by keeping Q = 4. This shifts the indexing of the previous local optima so that the previous x(2), x(3) and x(4) now become x(1), x(2) and x(3). The new EE1(j) and EE0(j) values may be verified by consulting the new vectors that now qualify as x(1) through x(4). Alternatively, these values can be computed from the inductive formula (5). (For example, in the case of $x_1$, which currently equals 1 in x(4), the new value for EE1(1) is 8 + .5·9 = 12.5, and rounding down with integer arithmetic gives EE1(1) = 12.)

This example brings up an additional characteristic of the method. The final Move 8 that gives $x_3$ an S2 status affords the simplest way to launch an Ascent Phase. Specifically, a variable $x_j$ with an S2 status that becomes the "last variable" to satisfy the trigger threshold already has received an evaluation Eval(j) ≤ 0 and is already tabu. Thus, no change is required in $x_j$ or its tabu status to launch a new ascent.

However, if a last variable to satisfy the trigger threshold does so by receiving an S1 status, then it would be necessary to make the move that gives $x_j$ its new value. (This could have happened in the Working Table if Move 8 had caused $x_8$ to qualify for an S1 status instead of causing $x_3$ to qualify for an S2 status.) Then, after giving $x_j$ its new value, its evaluation Eval(j) will be negated to yield Eval(j) < 0 and $x_j$ will be made tabu to launch the new ascent. This final move could cause StatusCount to drop if it cancels the S1 or S2 status of some other variable(s), but there is a simple way to handle this. Rather than keeping track of cancellations, we keep a value StatusCount1 for an S1 status and StatusCount2 for an S2 status. Then we only increase StatusCount1 by 1 when a variable $x_j$ with an S1 status is chosen to change its value (which locks in the value for $x_j$) and each time a variable changes its value we recompute StatusCount2 (starting over from StatusCount2 = 0). Then we identify the current StatusCount2 value at the same time as scanning the variables to select a new $x_j$ to change its value.

We call an algorithm that follows the design of this example to alternate between an Ascent Phase and a Post-Ascent Phase, by exploiting the recency threshold and the trigger threshold, an Alternating Ascent (AA) algorithm. Choice rules for selecting a variable $x_j$ to become $x_k$ in the AA algorithm are discussed in Section 5 followed by a pseudocode in Section 6 that provides an effective way to implement the algorithm.

## 5. Choice Rules

Customary choice rules to select a variable $x_j$ to become $x_k$ and change its value from $x_k = x_k^\#$ to $x_k = 1 - x_k^\#$ can be extended in the context of the AA algorithm to take advantage of the values EE(j) incorporated in the recency threshold. We identify three options for doing this, a simple weighting scheme, a simple cutoff (threshold) scheme and a more advanced cutoff procedure using an evaluation tradeoff analysis. The algorithm will only use one of these options, hence providing three different versions of the algorithm. These choice rule options depend on differentiating two conditions related to the criteria used to determine an S1 and S2 status.

To describe these conditions, we define an *iteration* to consist of examining each of the variables $x_j$, $j \in N$, to identify one to become the variable $x_k$ that is selected to change its value.

**Condition 1** *(Represented by Condition1 = True).*
Eval(j) > 0 and either $x_j$ is not tabu (TabuIter(j) < Iter) or satisfies the threshold criterion EE(j) ≥ Threshold(r). This condition is relevant to all iterations in an Ascent Phase and to iterations in a Post-Ascent Phase where some variable has an S+ or S1 status.

**Condition 2.** *(Represented by Condition2 = True).*
Eval(j) ≤ 0 and $x_j$ is not tabu (TabuIter(j) < Iter). This condition is relevant to iterations in a Post-Ascent Phase.

These conditions set the stage for different responses to select a variable $x_j$ to become the variable $x_k$ in the AA algorithm. To begin an iteration, both Condition1 = False and Condition2 = False, until the respective criteria for Conditions 1 and 2 are satisfied. Condition 1 takes priority, in that once a variable $x_j$ is examined that yields Condition1 = True, then the only variables $x_j$ that will be candidates for becoming $x_k$ on the current iteration are those that also satisfy the criteria for Condition 1, hence only variables with Eval(j) > 0 are considered thereafter.

As subsequently described, we provide two approaches for exploiting the information made available by Conditions 1 and 2, a Single-Pass AA algorithm and a Double-Pass AA algorithm. The Double-Pass algorithm operates by employing a First Pass that gathers information as a basis for making more judicious choices of variables to change their values during the Second Pass.

It is expected for most settings that the Double-Pass algorithm will provide better decisions than the Single-Pass Algorithm, but the Single-Pass algorithm has the advantage of requiring less computation during each iteration. We first examine the types of choice rule options for the Single-Pass algorithm, which require an analysis of information generated sequentially. (A compensating advantage of the Double-Pass approach is to avoid the intricacies of a sequential analysis.)

Let BestEval, BestEE. BestEvalW and MaxEE respectively identify the best Eval(j) value, the best EE(j) value, the best weighted Eval(j) value and the maximum EE(j) value found in the sequential examination of the variables $x_j$ during an iteration of the Single-Pass approach, where "best" refers to the values determined by the evaluation criteria used.

For all choice rules of the Single-Pass approach, we first apply a dominance check

$$\text{Eval}(j) \geq \text{BestEval and } \text{EE}(j) \geq \text{MaxEE}$$

If this dominance criterion is satisfied $x_j$ is automatically considered a candidate to become $x_k$, using the update BestEval = Eval(j) and MaxEE = EE(j), together with setting BestEE = EE(j) and k = j. If the dominance criterion is not satisfied, then the method uses one of the following three options.

**Simple Weighted Sum Rule**

Weight parameters W1 for Condition 1 and W2 for Condition 2 are used to modify the evaluation Eval(j), by setting W = W1 or W2 when Condition1 or Condition2 = True to produce the evaluation EvalW given by

$$\text{EvalW} = \text{Eval}(j) + W \cdot (\text{EE}(j)/\text{EEbase}).$$

The normalization of dividing EE(j) by EEbase gives $1 \geq \text{EE}(j)/\text{EEbase} \geq 0$, which makes the calibration of W easier.

Whenever EvalW > BestEvalW the method updates BestEvalW = EvalW, BestEval = Max(BestEval, Eval(j)) and MaxEE = Max(MaxEE, EE(j)), together with setting k = j. (If the dominance check were not used, it would not be necessary to update BestEval and MaxEE.) We always know whether BestEvalW refers to W = W1 or W2 because the Simple Weghted Sum Rule is implemented separately for Condition1 = True and Condition2 = True.

The weight value W = W1 producing EvalW for Condition 1 will generally be small, as in performing a tie-breaking function. The value W = W2 for Condition 2 may also be small, but intuition suggests it may preferably be larger, perhaps in some instances large enough to cause EE(j) to dominate Eval(j). The possibilities for both W1 and W2 may range, for example, from .1 to the maximum expected value for Eval(j).

**Simple and Advanced Cutoff Rules**

The remaining choice rule options are given by the Simple Cutoff Rule and the Advanced Cutoff Rule, and are preceded by checking a secondary dominance criterion

$$\text{EE}(j) \geq \text{BestEE and Eval}(j) \geq \text{BestEval}$$

If satisfied, this criterion results in updating BestEval = Eval(j), BestEE = EE(j) and MaxEE = Max(MaxEE, BestEE), together with setting k = j. When the secondary dominance criterion is not satisfied the Simple and Advanced Cutoff Rules (which are applied separately in different versions of the algorithm) are as follows.

**Simple Cutoff Rule**

The idea underlying this rule is to consider that the current $x_j$ is a better candidate for $x_k$ than previous candidates if

$$EE(j) \geq Cutoff \text{ and } Eval(j) > BestEval$$

where Cutoff = F·MaxEE and F is a fraction chosen between .5 and .9 (or more restrictively, between .7 and .9). In the special case where the S1 status applies, Cutoff = Max(F·MaxEE, ThresholdR).

**Advanced Cutoff Rule**

The Advanced Cutoff Rule is based on the same Cutoff value but uses a criterion to identify tradeoffs between EE(j) and Eval(j), expressed in

$$EE(j) \geq Cutoff \text{ and } EE(j) \cdot Eval(j) > BestEE \cdot BestEval$$

The analysis underlying the tradeoff inequality EE(j)·Eval(j) > BestEE·BestEval is explained in Appendix 4.

The preceding versions of the Simple and Advanced Cutoff Rules apply to Condition 1. Different versions apply to Condition 2, as elaborated in Section 7.

**Choice Rule for the Double-Pass AA algorithm**

The Double-Pass algorithm likewise uses a Cutoff Rule, but which is not applied until the Second Pass after accumulating information on the First Pass that provides the foundation for the rule. The information consists of identifying Min, Max and Mean values of Eval(j) and EE(j) on the current iteration and using these to compute cutoff values EvalCutoff and EECutoff that respectively provide alternative versions of the algorithm. The EvalCutoff version imposes the condition Eval(j) ≥ EvalCutoff and subject to this maximizes EE(j), while the EECutoff version imposes the condition EE(j) ≥ EECutoff and subject to this maximizes Eval(j). These Cutoff inequalities are effectively threshold inequalities that relate to the recency threshold EE(j) ≥ Threshold(r) of (4).

Details of these choice rules and their variations are given in the next two sections.

## 6. General AA Algorithm Design and Pseudocode

The tabu restrictions required during the Ascent Phase and the Post-Ascent Phase of the AA algorithm take a simple form where the tenure is given by setting Tenure = Large, where Large represents a large positive number. This simple approach is made possible by the fact that tabu restrictions will be overruled by the aspiration criterion and by an S1 status, which, together with the trigger threshold, implicitly determine the duration of a tabu restriction. Consequently, it would also be possible to represent the tabu restrictions by Tabu(j) = True or False, which would eliminate the parameter Tenure when setting TabuIter(j) = Iter + Tenure (which can now be replaced by Tabu(j) = True). However, we retain the TabuIter(j) representation to allow for applications where manipulating Tenure may prove useful.

Each iteration of the AA algorithm begins by checking the aspiration criterion for overriding a tabu restriction to see whether changing the value of $x_j$ will yield a value for $x_o$ (currently given by $x_o = x_o^\#$) that improves upon the best value $x_o^*$, as indicated by $x_o^\# + Eval(j) > x_o^*$. When $x_o^\# + Eval(j) > x_o^*$ is satisfied, the method updates $x_o^* = x_o^\# + Eval(j)$ and sets k = j along with setting Aspiration = True. By beginning the iteration in this fashion, it is unnecessary to examine $x_j$ further to see if $x_j$ is tabu or satisfies Condition 1 or Condition 2. Moreover, once Aspiration = True, each additional $x_j$ examined only needs to be checked to determine if it further improves $x_o$ by satisfying $x_o^\# + Eval(j) > x_o^*$. This use of Aspiration (which begins False until the aspiration criterion is satisfied), also has the function of choosing a best improving move in the simple case where no tabu moves exist, as during an initial ascent to reach a first local optimum.

As previously noted, the AA algorithm comes in two forms, a Single-Pass algorithm and a Double-Pass algorithm. While the Double-Pass algorithm is expected to be more effective, the reduced computation per iteration of the Single-Pass method motivates its inclusion as an option for consideration.

We first describe the Single-Pass algorithm and provide details of each of its main routines. Section 7, following, provides a description of its choice rules. The Double-Pass Algorithm is then described in Section 8.

**Note**: The comments in italics in the pseudocode for component routines are more extensive than customary to help convey the rationale of the processes employed. This is done to provide explanations within the context of these processes in contrast to offering explanations outside of these processes where the relevance of the explanations is less apparent.

**Single-Pass AA Algorithm**
    **Preliminary Initialization**
    For Iter = 1 to MaxIter
        *(or until the expiration of a time limit)*
        **Initialization for the Current Iteration**
        **Current Iteration Routine**
        **Post Iteration Update**
    Endfor
**End of the Single-Pass Algorithm**

The instructions for the four key components of the method – the Preliminary Initialization, the Initialization for the Current Iteration, the Current Iteration Routine and the Post Iteration Update – are as follows. We have included reference to instructions in the Preliminary Initialization and the Post Iteration Update that are to be inserted for the case of the Double-Pass Algorithm described in Section 8 and the Myopic Correction Method described in Section 9.

**Preliminary Initialization:**
*For the Weighted Sum Rule*
    W1 = .1 and W2 = 10 (e.g., where both may range from .01 to 100)

*For the Cutoff Rules*
    Choose the fraction F = .8 (e.g., F = .7 to .9)

*Values for general parameters:*
    Tenure = Large
    Small = 0 (or 1 or 2)
    Choose Q (e.g., Q = 15 to 25)
    Choose r < Q for Threshold(r) (e.g., r = 10 to 20)
    Choose Trigger (e.g., Trigger = 5 to 9)
    *(For simplicity, the value of Q may simply be chosen to be larger than any value expected for r, or real value arithmetic can be used in the inductive update of EE(j) and EEbase to avoid having $2^Q$ be an excessively large number. See Appendix 1 for further observations on choices of r, F and Trigger.)*

*Remaining Initialization:*
EEbase = $2^{Q-1}$ *(as in (5.1))*
Threshold$_o$ = Threshold(r) = $2^{Q-1} + 2^{Q-2} + \ldots + 2^{Q-r}$ *(as in (5.3))*
ThresholdR = Min(EEbase, Threshold$_o$) *(as in (5.4))*
For j = 1 to n
    TabuIter(j) = 0 *(No variables are tabu)*
    $x_j^\#$ = 0 *(Start with all $x_j$ = 0: See Note 1 below.)*
    Identify (compute) Eval(j) for the initial solution with all $x_j$ = 0
    EE(j) = EEbase *(= EE0(j) since $x_j^\#$ = 0)*

       *(Implicitly, EE1(j) = 0 and EE0(j) = EEbase. See Note 2 below.)*
Endfor
$x_o^\# = 0$

***Note 1:*** *For diversification approaches, $x_j$ can be initialized at other than $x_j = 0$ simply by complementing $x_j := 1 - x_j$, which then causes $x_j^\# = 0$ to correspond to $x_j = 1 - x_j^\# = 1$ for complemented variables.*

***Note 2****: EE1(j) and EE0(j) do not need to be recorded. It is sufficient to initialize EE(j) by EE(j) = EEbase, as above. Then the instructions of subsequent routines will update EE(j) appropriately.)*

*(Initialize the record of the 3 most recent variables $x_j$ assigned values in the Ascent Phase. This can be done for any number of recent variables assigned values in the Ascent Phase.)*
Recent1 = Recent2 = Recent3 = 0
Ascent = True *(Begin with an Ascent Phase)*
*Include for the Double-Pass Algorithm:* LastLink = n + 1
*Include for Myopic Correction Method (Section 9):*
       Select (AddNum, DropNum) schedule
       Insert MC1
**End Preliminary Initialization**

**Initialization for the Current Iteration**
StatusCount1 = 0
StatusCount2 = 0
Condition1 = False
Condition2 = False
S1 = False
Aspiration = False
*(The next four terms are used in the Choice Rule Options)*
BestEvalW = – Large
BestEval = – Large
BestEE = – Large
MaxEE = – Large
k = 0 *($x_k$ will be the variable chosen to change its value, and k will remain at k = 0 until a candidate variable $x_j$ to become $x_k$ is identified.)*
**End Initialization for the Current Iteration**

**Current Iteration Routine**
*Iteration Loop*
For j = 1 to n
    *Begin by checking the aspiration criterion. During an initial ascent, Aspiration = True will result at every move until no more improving moves exist.*

If $x_o^\# + Eval(j) > x_o^*$ then
   Aspiration = True
   k = j  *($x_k$ is chosen as the variable to change its value)*
   $x_o^* = x_o^\# + Eval(j)$
Endif
*(Once Aspiration = True, it remains True throughout the remainder of the iteration, and after checking whether $x_o^\# + Eval(j) > x_o^*$ for the current j, no additional check of j is needed.)*
If Aspiration = True
   Continue the Iteration Loop by going to the next j (or exiting the loop if j = n)
Endif
*(Reach here if Aspiration = False)*
If Condition1 = True *(Note the iteration loop begins with Condition1 = False. This can*
    *change to give Condition1 = True by checking under Condition1 = False, below.)*
    *(Given Condition1 = True, only consider Eval(j) > 0 for all subsequent j examined.)*
    If Eval(j) > 0 then
       If TabuIter(j) < Iter *($x_j$ is not tabu)* or EE(j) ≥ ThresholdR then
          *($x_j$ is admissible)*
          If S1 = False and EE(j) ≥ ThresholdR then
             *(S1 becomes True for the first time)*
             S1 = True
             *(Set BestEvalW, BestEval, BestEE, MaxEE as if starting from scratch.)*
             BestEvalW = Eval(j) + W1·EE(j)
             BestEval = Eval(j)
             BestEE = EE(j)
             MaxEE = EE(j)
             k = j
          Else *(S1 is not True)*
             *(Check for dominance)*
             If EE(j) ≥ MaxEE and Eval(j) ≥ BestEval then
                BestEvalW = Eval(j) + W1·EE(j)
                BestEval = Eval(j)
                BestEE = EE(j)
                MaxEE = EE(j)
                k = j
             Else
                *(The preceding dominance criterion does not hold. Apply one of the*
                *Choice Options 1.1, 1.2 or 1.3 for Condition 1, as subsequently*
                *described.)*
                **Execute Condition 1 Choice Routine**
             Endif
          Endif
       Endif

        Endif  
Else *(Condition1 = False – hence Condition1 has remained False from beginning to now.)*  
    If Eval(j) > 0 then  
        If TabuIter(j) < Iter or EE(j) $\geq$ Threshold(r) then  
            *($x_j$ is not tabu or the recency threshold is satisfied when Eval(j) > 0:*  
            *Change from Condition1 = False to Condition1 = True for the first time.)*  
            Condition1 = True  
            *(Set BestEvalW, BestEval, BestEE, MaxEE as if starting from scratch.)*  
            BestEvalW = Eval(j) + W1·EE(j)  
            BestEval = Eval(j)  
            BestEE = EE(j)  
            MaxEE = EE(j)  
            k = j  
            *(S1 then becomes True for the first time if the recency threshold holds)*  
            If EE(j) $\geq$ ThresholdR then S1 = True  
        Endif  
    Else *(Eval(j) $\leq$ 0 and Condition 1 remains False)*  
        *(Ascent = False makes it possible to choose a variable with Eval(j) $\leq$ 0. When*  
        *Ascent = True, it is only acceptable to choose a variable with Eval(j) > 0)*  
        If Ascent = False then  
            If EE(j) $\leq$ EEbase – ThresholdR then  
                *(This inequality implies $x_j = 1 - x_j^{\#}$ satisfies the recency threshold, and*  
                *the current value $x_j = x_j^{\#}$ does not equal $x_j^Q$, the value $x_j$ had in the*  
                *most recent local opt. Hence $x_j$ qualifies for S2 status since Eval(j) $\leq$ 0.)*  
                StatusCount2 := StatusCount2 + 1  
                LastVbl = j *($x_j$ is the last variable identified with S2 status)*  
            Elseif TabuIter(j) < Iter *($x_j$ is not tabu)* then  
                Condition2 = True *(for information only, not used for decision)*  
                *(Evaluate $x_j$ as a candidate for $x_k$ under Condition 2)*  
                *(Check for dominance)*  
                If EE(j) $\geq$ MaxEE and Eval(j) $\geq$ BestEval  
                    BestEvalW = Eval(j) + W2·EE(j)  
                    BestEval = Eval(j)  
                    BestEE = EE(j)  
                    MaxEE = EE(j)  
                    k = j  
                Else  
                  *(The preceding dominance criterion does not hold. Apply one*  
                  *of the Choice Options 2.1, 2.2 or 2.3 for Condition 2, as subsequently*  
                  *described.)*  
                  **Execute Condition 2 Choice Routine**  
                Endif  
            Endif

    Endif
   Endif
  Endif
Endfor
*End of Iteration Loop*
**End of Current Iteration**

Details of the Condition 1 and Condition 2 Choice Routines are given in Section 7, following.

**Post Iteration Update**
If k > 0 (*A variable $x_k$ was chosen to change its value*) then
 *(The method does not reach a new local optimum here but may launch a*
 *new Ascent Phase if StatusCount1 + StatusCount2) ≥ Trigger, which can only happen*
 *when Ascent = False. Condition1 = False implies a profitable non-tabu $x_j$ was never found,*
 *so if Condition1 = False and the Trigger threshold are both satisfied, should launch*
 *Ascent without setting $x_k$.)*
 If Condition1 = False and StatusCount1 + StatusCount2 ≥ Trigger then
  ***(Launch a new Ascent Phase)***
  Ascent = True
  *For Myopic Correction Method:* Insert MC1
  *(free all variables from tabu restrictions, except for LastVbl)*
  TabuIter(j) = 0 for j = 1 to n
  TabuIter(LastVbl) = Large.
  ***Exit the Post Iteration Update***
 Endif
 *(New ascent is not yet launched, but may be below)*
 *(update the current value $x_k^\#$ of $x_k$ and $x_o^\#$ of $x_o$)*
 $x_k^\# := 1 - x_k^\#$
 $x_o^\# := x_o^\# + Eval(k)$
 *(Update EE(k), which is the only EE(j) to update when $x_k$ changes its value)*
 EE(k) := EEbase – EE(k)

 If Ascent = True then
  *(save the 3 most recent variables that have changed their values, to set them tabu once*
  *the ascent ends. Recent1 = the most recent, Recent2 = the next most recent, etc. The*
  *same pattern can be used to save more or less than 3 variables.)*
  Recent3 = Recent2;  Recent2 = Recent1;  Recent1 = k
 Else *(Ascent = False and the search is in the Post-Ascent Phase)*
  *(make $x_k$ tabu if and only if Ascent = False)*
  TabuIter(k) = Iter + Tenure
  If Eval(k) > 0 then
   LastVbl = k *(to record the most recent $x_k$ set for Eval(k) > 0. This will be the last*
    *$x_k$ set by Condition 1 for checking if the Trigger threshold is satisfied to*

    *launch a new ascent.)*
   StatusCount2 = 0 *(all S2 conditions are removed when a move is made for*
    *Eval(k) > 0.)*
  Endif
  If Aspiration = True or S1 = True then StatusCount1 := StatusCount1 + 1
  If StatusCount1 + StatusCount2 ≥ Trigger then
   ***(Launch a new Ascent Phase)***
   *For Myopic Correction Method:* Insert MC1
   Ascent = True
   *(free all variables from tabu restrictions, except for LastVbl)*
   TabuIter(j) = 0 for j = 1 to n
   TabuIter(LastVbl) = Large.
  Endif
 Endif
 *(The following Eval(j) update is done here instead of immediately after changing $x_k^\#$ to permit checking Eval(k) > 0 above.)*
 Update Eval(j) for all j = 1 to n
 *For Myopic Correction Method:* Insert MC2
Else *(k = 0)*
 *(No variable $x_k$ could be chosen to change its value, hence must end an Ascent Phase if Ascent = True or must begin an Ascent Phase if Ascent = False. The outcome k = 0 is the only way to end an Ascent Phase. Eval(j) does not need to be updated.)*
 If Ascent = True then
  *(End an ascent in two steps: first, free LastVbl from its tabu restriction to complete the ascent to a local optimum, and then at the local optimum perform updates for the Post-Ascent Phase when Ascent = False.)*
  If LastVbl > 0 then
   ***(A conditional local optimum is reached)***
   *(free $x_j$ from its tabu restriction for j = LastVbl)*
   TabuIter(LastVbl) = 0
   *(now all variables are not tabu)*
   LastVbl = 0
  Else *(LastVbl = 0)*
   ***(A true local optimum is reached, and a Post-Ascent Phase begins)***
   *For Myopic Correction Method:* Insert MC1
   Ascent = False
   *(Update EEbase by (5.2) and ThresholdR by (5.4))*
   EEbase := $2^{Q-1}$ + EEbase/2
   ThresholdR = Min(EEbase, Threshold$_o$)
   For j = 1 to n
    If $x_j^\#$ = 1 then
     EE(j) := $2^{Q-1}$ + EE(j)/2
    Else *($x_j^\#$ = 0)*

$$EE(j) := (EEbase + EE(j))/2 - 2^{Q-1}$$
     Endif
    Endfor

    *Note: The preceding "for loop" is equivalent to the longer calculation of the following update based on saving the EE1(j) and EE0(j) values separately to determine EE(j):*
    For j = 1 to n
      $EE1(j) = 2^{Q-1}x_j^{\#} + EE1(j)/2$
      $EE0(j) = EEbase - EE1(j)$
      If $x_j^{\#} = 1$ then
        $EE(j) = EE1(j)$
      Else ($x_j^{\#} = 0$)
        $EE(j) = EE0(j)$
      Endif
    Endfor

    *(Make the most recent assigned variables tabu: By allowing TabuIter(0) to be a dummy TabuIter(j) value for j = 0, there is no need to check if j > 0)*
    TabuIter(j) = Iter + Tenure for j = Recent1, Recent2 and Recent3.
   Endif
  Else *(Ascent = False with no admissible $x_k$ found to change its value)*
   *(**Launch a new Ascent Phase**)*
   Ascent = True
   *(free all variables from tabu restrictions, except for LastVbl)*
   TabuIter(j) = 0 for j = 1 to n
   TabuIter(LastVbl) = Iter + Tenure. *(Option: let LastVbl = 0 as a signal to skip setting $x_j$ tabu for j = LastVbl > 0, so that all variables are not tabu.)*
   *(Initialize the 3 most recent variables assigned values for the Ascent Phase)*
   Recent1 = Recent2 = Recent3 = 0
  Endif
Endif
**End of Post Iteration Update**

## 7. Choice Rule Routines for the Single-Pass AA Algorithm

We now identify the three main options for each choice rule of the Single-Pass algorithm, consisting of the Weighted Sum Rule, the Simple Cutoff Rule and the Advanced Cutoff Rule.

As previously noted, each of these choice rule options gives rise to a different version of the algorithm.

**Condition 1 Choice Routine**
*(Apply one of the three Choice Options 1.1, 1.2 or 1.3 for Condition 1, as follows.)*

---

      *Choice Option 1.1: Weighted Sum Rule*
If Eval(j) + W1·EE(j) > BestEvalW then
    BestEvalW = Eval(j) + W1·EE(j)
    BestEval = Max(BestEval, Eval(j))
    MaxEE = Max(MaxEE, EE(j))
    k = j
Endif

---

The Choice Options 1.2 and 1.3 are preceded by the following secondary dominance check.

    *(Secondary dominance check)*
If EE(j) ≥ BestEE and Eval(j) ≥ BestEval then
    BestEval = Eval(j)
    BestEE = EE(j)
    MaxEE = Max(MaxEE, BestEE)
    k = j
Else *(secondary dominance does not hold)*
    *(Identify Cutoff to prepare for Choice Rules)*
    Cutoff = F·MaxEE
    If S1 = True then Cutoff = Max(Cutoff, ThresholdR)

---

      **Choice Option 1.2: Simple Cutoff Rule**
If EE(j) ≥ Cutoff and Eval(j) > BestEval then
    BestEval = Eval(j)
    BestEE = EE(j)
    MaxEE = Max(MaxEE, BestEE)
    k = j
Endif

---

      **Choice Option 1.3: Advanced Cutoff Rule**
If EE(j) ≥ Cutoff and EE(j)·Eval(j) > BestEE·BestEval then
    BestEval = Eval(j)
    BestEE = EE(j)
    MaxEE = Max(MaxEE, BestEE)

          k = j
     Endif
*(Note: It is possible for Option 1.3 to rely more fully on the tradeoffs in the Inequality EE(j)·Eval(j) > BestEE·BestEval by choosing F smaller, for example $F \leq .5$, and not using ThresholdR to define Cutoff.)*

-----------------------------------------------------------------------------

  Endif (in the Current Iteration loop)

**Condition 2 Choice Routine**
*(The dominance criterion does not hold. Apply one of the three Choice Options 2.1, 2.2 or 2.3 for Condition 2, as follows.)*

-----------------------------------------------------------------------------

          **Choice Option 2.1: Weighted Sum Rule**
  If Eval(j) + W2·EE(j) > BestEvalW then
     BestEvalW = Eval(j) + W2·EE(j)
     BestEval = Max(BestEval, Eval(j))
     MaxEE = Max(MaxEE, EE(j))
     k = j

-----------------------------------------------------------------------------

The Choice Options 2.2 and 2.3 are preceded by checking a secondary dominance criterion, as follows.

     *(Secondary dominance check)*
  If EE(j) ≥ BestEE and Eval(j) ≥ BestEval then
     BestEval = Eval(j)
     BestEE = EE(j)
     MaxEE = Max(MaxEE, BestEE)
     k = j
  Else *(secondary dominance does not hold)*
     *(Identify Cutoff to prepare for Choice Rules)*
     Cutoff = (1/F)·MaxEE

-----------------------------------------------------------------------------

          **Choice Option 2.2: Simple Cutoff Rule**
    If EE(j) ≥ Cutoff and Eval(j) > BestEval then
       BestEval = Eval(j)
       BestEE = EE(j)
       MaxEE = Max(MaxEE, BestEE)
       k = j

       Endif

--------------------------------------------------------------------------------

       **Choice Option 2.3: Advanced Cutoff Rule**
    If EE(j) ≥ Cutoff and Eval(j)·BestEE > BestEval·EE(j) then
        BestEval = Eval(j)
        BestEE = EE(j)
        MaxEE = Max(MaxEE, BestEE)
        k = j
    Endif
*(Note: It is possible for Option 2.3 to rely more fully on the tradeoffs in the Inequality Eval(j)·BestEE > BestEval·EE(j) by choosing F larger, for example F > .9.)*

--------------------------------------------------------------------------------

   Endif (in the Current Iteration loop)

It may be noted that the difference between the Cutoff Rule Options for Conditions 1 and 2 lies in the definition of Cutoff (which additionally refers to ThresholdR in Condition 1 and replaces F by 1/F in Condition 2) and in the difference between the tradeoff inequalities (Eval(j)·BestEval > EE(j)·BestEE in Condition 1 and Eval(j)·BestEE > BestEval·EE(j) in Condition 2).

Each of the Cutoff Rule Options 1.2, 1.3, 2.2 and 2.3 has a counterpart that reverses the role of EE and Eval. The counterparts result by keeping track of MaxEval instead of MaxEE – by updating MaxEval = Max(MaxEval, BestEval), and replacing MaxEE by MaxEval in the definition of Cutoff – and may be viewed as sequential approximations of the EvalCutoff threshold rule of the next section.

## 8. The Double-Pass AA Algorithm

As noted in Section 5, the Double-Pass algorithm comes in two versions, according to whether the algorithm uses a cutoff EvalCutoff for Eval(j), or a cutoff EECutoff for EE(j).

Each cutoff is a lower limit computed from the Min, Max and Mean values for Eval(j) or EE(j) that allows only the "Top Percent" (20%, 10% or 5%, etc.) of the variables to satisfy the cutoff criterion

$$\text{Eval}(j) \geq \text{EvalCutoff} \tag{A1}$$

or

$$\text{EE}(j) \geq \text{EECutoff} \tag{A2}$$

The EvalCutoff choice rule then selects an admissible variable $x_k$ that maximizes EE(j) subject to (A1) and the EECutoff choice rule selects an admissible variable $x_k$ that maximizes Eval(j) subject to (A2).

When Condition1 = True, we restrict attention to variables $x_j$ with Eval(j) > 0 in considering variables that satisfy either (A1) or (A2), and similarly, when S1 = True, we restrict attention to variables that additionally satisfy the recency threshold EE(j) ≥ ThresholdR. We observe that (A1) and (A2) may be viewed as "variable" threshold inequalities whose right-hand sides depend on the information of the current iteration.

The First Pass of the Double-Pass method thus determines whether Condition 1 or Condition 2 applies, and simultaneously determines the values used to compute EvalCutoff and EECutoff used in the Second Pass.

To facilitate the execution of the Second Pass, we introduce a linked list Link(j) to record the indexes j of those variables $x_j$ that are identified on the First Pass as candidates to be selected as the variable $x_k$. The use of this linked list saves effort in the Second Pass by disregarding all variables that do not qualify as candidates to be chosen. The reduction in the number of $x_j$ variables examined in the Second Pass can be substantial when Condition1 = True, and more substantial when, in addition, S1 = True.

The linked list Link(j) starts by saving the first candidate encountered by setting FirstLink = j and Link(j) = LastLink (where LastLink = n + 1 to differentiate it from the variables indexed from j = 1 to n). Then each new candidate $x_j$ encountered is added to the list by setting Link(j) = FirstLink and FirstLink = j. Consequently, all candidates recorded on the list can be recovered by starting with j = FirstLink and successively setting j := Link(j) until reaching j = LastLink.

The first $x_j$ encountered that yields Condition1 = True reinitiates the linked list memory with the starting assignment Link(j) = FirstLink and FirstLink = j, since from this point onward the only candidates for $x_k$ will be those that qualify by satisfying Condition 1. The starting assignment is likewise reinitiated for the first $x_j$ encountered that yields S1 = True, for the same reason.

The following code is organized to choose either to exclude the use of the linked list (to save memory) and examine all j indexes on the Second Pass, or to include the linked list and examine only the qualifying indexes j on the Second Pass.

We describe the Double-Pass algorithm by presenting the EvalCutoff and the EECutoff versions together, to show their similarities and differences. We have partitioned the instructions for the two versions so that each can easily be implemented separately from the other.

At the risk of redundancy, italicized comments are liberally used, as before, to help explain the processes represented by the pseudocode.

**Double-Pass AA Algorithm**
    **Preliminary Initialization**
    For Iter = 1 to MaxIter
        *(or until the expiration of a time limit)*
        **Initialization for the Current Iteration**
        **Current Iteration Routine (encompassing the First Pass and Second Pass)**
        **Post Iteration Update**
    Endfor
**End of the Double-Pass Algorithm**

The Current Iteration Routine incorporates the iteration routines for the First Pass and Second Pass Routines, based on the values k and Aspiration produced by the First Pass, as follows.

**Current Iteration Routine**
    Execute the **First Pass Iteration Routine**
    If $k > 0$ and Aspiration = False then
        Execute the **Second Pass Iteration Routine**
    Endif

The Preliminary Initialization (except for setting parameters for the Weighted Choice Rule), and the Post Iteration Update are as given in the Single-Pass algorithm. The Initialization for the Current Iteration and the Iteration Routines for the First and Second Passes are as follows.

**Initialization for Current Iteration**
*(Applicable to both the First and Second Passes)*
StatusCount1 = 0
StatusCount2 = 0
Condition1 = False
Condition2 = False
S1 = False
Aspiration = False
jFirst = 0
jLast = 0
---------------------
*Version 1: Eval Cutoff Version – each iteration maximizes EE(j) subject to Eval(j) $\geq$ EvalCutoff*
MaxEval = – Large
MinEval = Large
SumEval = 0
CountEval = 0
    *(SumEval and CountEval will be used to compute MeanEval = SumEval/CountEval)*
---------------------
*Version 2: EE Cutoff Version – each iteration maximizes Eval(j) subject to EE(j) $\geq$ EECutoff*
MaxEE = – Large

MinEE = Large
SumEE = 0
CountEE = 0
    *(SumEE and CountEE will be used to compute MeanEE = SumEE/CountEE)*
-----------------------
k = 0
*(k is used as a flag in the First Pass to indicate whether a variable $x_j$ exists as a candidate for $x_k$ and is used in the Second Pass to identify the variable chosen to change its value. Exception: If the first Pass sets Aspiration = True, then k refers to the chosen variable $x_k$ and the Iteration Routine for the Second Pass is skipped, as indicated in the code above that gives the form of the Current Iteration Routine.)*
***End Initialization for the Current Iteration***

**First Pass Iteration Routine**
*Iteration Loop for First Pass*
For j = 1 to n
    *Begin by checking the aspiration criterion. During an initial ascent, Aspiration = True will result at every move until no more improving moves exist.*
    If $x_o^\#$ + Eval(j) > $x_o^*$ then
       Aspiration = True
       k = j  *($x_k$ is chosen as the variable to change its value)*
       $x_o^*$ = $x_o^\#$ + Eval(j)
    Endif
    *(Once Aspiration = True, it remains True throughout the remainder of the iteration, and after checking whether $x_o^\#$ + Eval(j) > $x_o^*$ for the current j, no additional check of j is needed.)*
    If Aspiration = True
       Continue the Iteration Loop by going to the next j (or exiting the loop if j = n)
    Endif
    *(Reach here if Aspiration = False)*
    If Condition1 = True *(The iteration loop begins with Condition1 = False, which can*
        *change to give Condition1 = True when checking under Condition1 = False, below.)*
        *(Given Condition1 = True, only consider Eval(j) > 0 for all subsequent j examined.)*
       If Eval(j) > 0 then
          If TabuIter(j) < Iter *($x_j$ is not tabu)* or EE(j) ≥ ThresholdR then
            *($x_j$ is admissible)*
            If S1 = False and EE(j) ≥ ThresholdR then
               *(S1 becomes True for the first time)*
               S1 = True
               *(Set Cutoff Info for Second Pass as if starting from scratch.)*
               Reinitiate jFirst and jLast
               jFirst = jLast = j

k = j *(k > 0 is a flag at the end of the First Pass that an admissible choice was possible, so that k = 0 implies no admissible choice exists, as in the Single-Pass method.)*

---
*Version 1: Reinitiate values for computing EvalCutoff*
MinEval = MaxEval = SumEval = Eval(j)
CountEval = 1

---
*Version 2 Reinitiate values for computing EECutoff*
MinEE = MaxEE= SumEE = EE(j)
CountEE = 1

---
*Reinitiate linked list*
FirstLink = j; Link(j) = LastLink

Else *(S1 is not True but Condition1 = True)*

---
*Version 1: Update values for computing EvalCutoff*
MinEval = Min(MinEval, Eval(j))
MaxEval = Max(MaxEval, Eval(j))
SumEval = SumEval + Eval(j)
CountEval := CountEval + 1

---
*Version 2: Update values for computing EECutoff*
MinEE = Min(MinEE, EE(j))
MaxEE = Max(MaxEE, EE(j))
SumEE = SumEE + EE(j)
CountEE := CountEE + 1

---
*Update jLast*
jLast = j
*Update linked list*
Link(j) = FirstLink; FirstLink = j
Endif
Endif
Endif
Else *(Condition1 = False – hence Condition1 has remained False from beginning to now.)*
If Eval(j) > 0 then
If TabuIter(j) < Iter or EE(j) ≥ Threshold(r) then
*($x_j$ is not tabu or the recency threshold is satisfied when Eval(j) > 0:*
Change from *Condition1 = False to Condition1 = True for the first time.)*
Condition1 = True
*(Set Cutoff Info as if starting from scratch.)*

*Reinitiate jFirst and jLast*
jFirst = jLast = j
k = j *(k > 0 is a flag at the end of the First Pass that an admissible choice was possible, so that k = 0 implies no admissible choice exists, as in the Single-Pass method.)*
-------------------------------------------------------------------------------------
*Version 1: Reinitiate values for computing EvalCutoff*
MinEval = MaxEval = SumEval = Eval(j)
CountEval = 1
-------------------------------------------------------------------------------------
*Version 2: Reinitiate values for computing EECutoff*
MinEE = MaxEE = SumEE = EE(j)
CountEE = 1
-------------------------------------------------------------------------------------
*Reinitiate linked list*
FirstLink = j; Link(j) = LastLink

*(S1 becomes True for the first time if the recency threshold holds)*
If EE(j) ≥ ThresholdR then S1 = True
    Endif
Else *(Eval(j) ≤ 0 and Condition 1 remains False)*
    *(Ascent = False makes it possible to choose a variable with Eval(j) ≤ 0. When Ascent = True, it is only acceptable to choose a variable with Eval(j) > 0)*
    If Ascent = False then
        If EE(j) ≤ EEbase – ThresholdR then
            *(This inequality implies $x_j = 1 - x_j^\#$ satisfies the recency threshold, and the current value $x_j = x_j^\#$ does not equal $x_j^Q$, the value $x_j$ received in the most recent local opt. Hence $x_j$ qualifies for S2 status since Eval(j) ≤ 0. Also know $x_j$ is tabu and $x_j$ does not qualify to be $x_k$.)*
            StatusCount2 := StatusCount2 + 1
            LastVbl = j *($x_j$ is the last variable identified with S2 status)*
        Elseif TabuIter(j) < Iter *($x_j$ is not tabu)* then
            Condition2 = True  *(for information only, not used for decision)*
            k = j *(k > 0 is a flag at the end of the First Pass that an admissible choice was possible, so that k = 0 implies no admissible choice exists, as in the Single-Pass method.)*
            *(Initialization previously set jFirst = 0.)*
            If jFirst = 0 then
                *Initiate jFirst and jLast*
                jFirst = j
                jLast = j
                *Initiate linked list*
                FirstLink = j and Link(j) = LastLink

                    Else  
                        *Update jLast*  
                        jLast = j  
                        *Update linked list*  
                        Link(j) = FirstLink; FirstLink = j  
                  Endif  

---

                  *Version 1: Update or initiate values for computing EvalCutoff*  
                  *(Initialization previously set MinEval = Large, MaxEval = − Large,*  
                      *SumEval = CountEval = 0, hence updating also initiates when*  
                      *Condition2 = True.)*  
                  MinEval = Min(MinEval, Eval(j))  
                  MaxEval = Max(MaxEval, Eval(j))  
                  SumEval = SumEval + Eval(j)  
                  CountEval := CountEval + 1  

---

                  *Version 2: Update or initiate values for computing EECutoff*  
                  *(Initialization previously set MinEE = Large, MaxEE = − Large,*  
                      *SumEE = CountEE = 0, hence updating also initiates when*  
                      *Condition2 = True.)*  
                  MinEE = Min(MinEE, EE(j))  
                  MaxEE = Max(MaxEE, EE(j))  
                  SumEE = SumEE + EE(j)  
                  CountEE := CountEE + 1  

---

                Endif  
              Endif  
            Endif  
          Endif  
Endfor  
*End of Iteration Loop*  
**End of First Pass Iteration Routine**

**Second Pass Iteration Routine**  
*(First prepare for the iteration loop of the Second Pass:)*  
MeanEval := SumEval/CountEval  
MeanEE := SumEE/CountEE  

*Calculate EvalCutoff and EECutoff by interpolation to approximately correspond to the fraction F of total number of $x_j$ that qualify for selection. F is treated as a percentile value (e.g., where F = .9 corresponds to the 90$^{th}$ percentile). The approximation is based on letting MeanEval correspond to F = .5, MaxEval correspond to F = 1.0, and MinEval correspond to F = 0.*

-----------------------------------------------------------------------------------------
*For Version 1:*
If F ≥ .5 then
       EvalCutoff = MeanEval + 2·(F − .5)(MaxEval − MeanEval)
Else
       EvalCutoff = MinEval + 2F·(MeanEval − MinEval)
Endif
-----------------------------------------------------------------------------------------
*For Version 2:*
If F ≥ .5 then
       EECutoff = MeanEE + 2·(F − .5)(MaxEE − MeanEE
Else
       EECutoff = MinEE + 2F·(MeanEE − MinEE)
Endif
-----------------------------------------------------------------------------------------
BestEE = BestEval = 0
k = 0
*If use the linked list:*
j = FirstLink

*(The aspiration criterion has already been checked on the First Pass and Aspiration = False when executing the Second Pass. The brief instructions for the Second Pass Iteration Loop are as follows.)*

*Second Pass Iteration Loop using the linked list*
While j < LastLink
    -------------------------------------------------------------------------------
    *For Version 1: using EvalCutoff*
    If Eval(j) ≥ EvalCutoff and EE(j) ≥ BestEE then
         BestEE = EE(j)
         k = j
    Endif
    -------------------------------------------------------------------------------
    *For Version 2: using EECutoff*
    If EE(j) ≥ EECutoff and Eval(j) ≥ BestEval then
         BestEval = Eval(j)
         k = j
    Endif
    -------------------------------------------------------------------------------
    j = Link(j)
Endwhile
*End of Second Pass Iteration loop using the linked list*

The iteration loop instructions for Versions 1 and 2 when not using the linked list are identical to those when using the linked list except for the larger range of j indexes examined.

*Second Pass Iteration loop without using the linked list*
For j = jFirst to jLast
    --------------------------------------------------------------------------------
    *For Version 1: using EvalCutoff*
    If Eval(j) ≥ EvalCutoff and EE(j) ≥ BestEE then
        BestEE = EE(j)
        k = j
    Endif
    --------------------------------------------------------------------------------
    *For Version 2: using EECutoff*
    If EE(j) ≥ EECutoff and Eval(j) ≥ BestEval then
        BestEval = Eval(j)
        k = j
    Endif
    --------------------------------------------------------------------------------
Endfor
*End of Second Pass Iteration loop without using the linked list*
**End of Second Pass Iteration Routine**

When the instructions for the unused cutoff version are removed, the size of the foregoing routines becomes appreciably smaller.

## 9. A Myopic Correction Strategy

The Myopic Correction (M-C) strategy provides a way to refine the AA algorithm to improve the trajectories of the Ascent and Post-Ascent phases. It is designed to counter two types of myopic bias identified in [3]:

*(Local Myopic Bias)* Early moves in a search process are more likely to be bad ones.

*(Reinforced Myopic Bias)* Early moves in a search process are likely to look better to retain – i.e., reversing them is likely to appear less attractive than they should – once later moves have been made.

These biases are commonly encountered by search processes in combinatorial optimization. As a rule, early moves are made with little information about their effect since not enough previous moves have accumulated to provide a foundation for evaluating current choices. Consequently, the early moves have a reasonable chance of being bad ones. Yet once made, these moves will influence the choice of moves that follow. The subsequent moves will therefore tend to reinforce the previous choices, making their earlier (faulty) evaluations appear to be justified. Such effects evidently depend on the region of the solution space where the search currently takes place, but without knowledge of whether this region lies in the vicinity of a global optimum, prudence suggests that it can be useful to guard against the possibility that these myopic effects may be operating. The goal is to identify a strategy for protecting against myopia – or rather, for correcting it by removing bad choices – that does not interfere with the capacity to make and retain good choices.

A potential response for combatting these myopic tendencies, which is the foundation of the Myopic Correction Method, is to periodically allow earlier moves to be reversed. An example of such a periodic reversal process occurs is as follows. Suppose that after making the first three moves, the first move is reversed (implicitly "dropped"), as if the move had not been made. Once dropped, the move's influence on the moves that follow will be diminished. (The first move will retain some level of influence, since the two moves that followed are still retained.) Continuing the process, after making a few more moves, the second move may be reversed, reducing its influence on subsequent moves. These "dropped" moves are given a tabu tenure of 0 or at most 1 or 2, so that they will immediately or almost immediately be free to be selected again. Consequently, if the decision to reverse the value of a variable $x_k$ was a mistake, as verified by the new evaluations after reversing $x_k$'s value (which may suggest $x_k$ should be returned to its previous value), then the previous value can be recovered at once and the search can proceed as if the reversal had not occurred. On the other hand, if the new evaluations disclose that a variable other than $x_k$ should change its value instead, this suggests that the reversal was appropriate.

Motivated by these considerations, we apply the following approach.

**The Myopic Correction Method**

A parameter AddNum is introduced that determines the number of variables that are "added" to the solution (by changing their values) before identifying variables to be "dropped," whose number is given by DropNum. Thus, after every AddNum steps of adding variables, DropNum steps are executed that drop variables. These two sequences of steps are called Add Steps and Drop Steps.

The alternation between Add Steps and Drop Steps continues until the search transitions from an Ascent Phase to a Post-Ascent Phase, or from a Post-Ascent Phase to an Ascent Phase, at which point the Myopic Correction Method is reinitialized and starts over again.

AddNum may start small at 2 or 3, and gradually increase to a value of 5 or 7. DropNum will be somewhat smaller than AddNum, and typically be limited to 1 or 2. (As always, such comments are suggestions that may be superseded.)

For a dropped variable $x_k$, the tabu tenure value can be given by Tenure = Small, where Small lies between 0 and 2. (A choice of Small = 0 is reasonable for preliminary experimentation.) The purpose of keeping the tenure of a dropped variable small is to leave $x_k$ free to change back to its previous value at once, or almost at once.

To illustrate, two possible schedules for selecting AddNum and DropNum are as follows.
- (A) Begin with (AddNum, DropNum) = (3, 1) and then after each Drop Step change (AddNum, DropNum) successively to the new values (4, 2), (5, 2), …, (5,2), (5,1), …, (5,1).
- (B) More restrictively, to make fewer changes on Drop Steps compared to Add Steps, begin with (AddNum, DropNum) = (3,1) and continue with (4, 1), (5,1), …, (5,1), (6,1), …, (6,1).

The more restrictive approach (B) is simpler and recommended for initial experimentation because it only drops a single variable on each Drop Step.

A counter denoted by MovesAdded starts at 0 and keeps track of the number of times that variables are successively selected and added without performing corrections that drop oldest variables. Each time a choice is made to add a variable, MovesAdded := MovesAdded + 1 until reaching the value AddNum. Then DropNum variables are identified to be reversed and dropped, accompanied by setting MovesAdded back to 0. Since variables change their values both during an Add Step and a Drop Step, the evaluations Eval(j) for all variables $x_j$ must be updated on each occasion.

We identify a linked list data structure for handling these processes that partly resembles the linked list approach used in the Double-Pass Algorithm, though its purpose and function is different.

**A Linked List Data Structure for the MC Method.**

A simple version of the linked list structure would let After(k) denote the linked list for recording variables $x_k$ as they change their values, where $x_k$ (the index k) is added to the end of the After list if $x_k$ changes its value by an Add Step (a choice step) and is removed from the start of the After list if $x_k$ changes its value by a Drop Step. However, this simple version encounters a difficulty when a variable $x_k$ that is already on the list, because of previously being selected to change its value, is selected again to change its value. Then the index k' = After(k) for the previous value of After(k) will be replaced by an index k" = After(k) for the new After(k) value, which will destroy access to k', preventing the After(k) list from being traced appropriately. One remedy is to create a double linked list structure, Before(k) and After(k), so that whenever $x_k$ is currently on the list and chosen to change its value, the current $x_k$ record can be overwritten by linking Before(k) to After(k), which permits the list to be traced as desired. However, this requires more steps and increases the memory, and additionally has the problem that removing the earlier record of $x_k$ causes the trace to skip over the old position of $x_k$ and drop a variable that should not be dropped. In other words, when the time comes to drop $x_k$ from the start of the list, $x_k$ could already have been moved from this position and will not be found there, causing the next variable in line to be dropped. Since $x_k$ has already been reversed, the step now reverses an additional variable (or more, if the next variable has also been moved). The following method overcomes this difficulty and uses less memory than the double linked list approach.

We take advantage of the fact that the number of moves made in an Ascent Phase or a Post-Ascent Phase will typically be far fewer than the number n of problem variables, particularly when n is relatively large. For example, if n is as great as several thousand, it would be extremely unlikely that every $x_j$ variable would change its value in one of these phases.

The After linked list array is structured so that instead of referring to After(k), we refer to After(h), where h is an index taken from the set H = {1, …, $h_o$ }. Supporting this, we identify k = ID(h) (ID for "identity") and h =RevID(k) (RevID for "reverse ID"). As shown below, the dimension $h_o$ of the arrays H, After and ID can be typically be chosen to be n/12 or smaller for problems where n is large, thus saving memory compared to using a linked list based directly on the indexes of the variables. The array RevID(k) ranges over all variables $x_k$, but by replacing After(k) with After(h) the total memory is less than using Before(k) and After(k), and the number of operations performed on each Add Step and Drop Step is also slightly smaller. Finally, using After(h) instead of After(k) we avoid the difficulty of losing access to a variable $x_k$ that has been moved.

The approach works as follows. H is recorded as a vector H(1), …, H(hLast), where hLast is initialized by setting hLast = $h_o$. The After list has a beginning dummy link BeginLink = 0 and an ending link EndLink that is initialized by setting EndLink = BeginLink together with After(BeginLink) = EndLink.

Each time a variable $x_k$ is chosen to change its value in an Add Step, the index h is added to the end of the After list by the sequence of instructions h = H(hLast), ID(h) = k, RevID(k) = h, After(EndLink) = h, EndLink = h. The Add Step then concludes by setting hLast := hLast – 1, so that a different h = H(hLast) will be added to the After list the next time the Add Step occurs. This update is skipped if hLast = 0 (meaning there are no more elements available on H to add to the After list).

Each time a variable is removed from the start of the After list by a Drop Step, the oldest variable $x_k$ accessed by the list is identified by h = After(BeginLink) and k = ID(h). If now h = RevID(k), we know that h identifies the location on the After list where $x_k$ most recently changed its value, which implies that the value of $x_k$ should be reversed. Otherwise, if h ≠ RevID(k), we know that $x_k$ has already reversed its value at least once since k was recorded by ID(h) = k and reversing $x_k$ now is inappropriate.

Regardless of whether $x_k$ reverses its value, we remove access to $x_k$ from location h at the start of the list by setting After(BeginLink) = After(h) and return h to the H vector by setting hLast :=

hLast + 1 and H(hLast) = h. The values RevID(k) and ID(h) need not be changed, since they will be modified appropriately when h is again removed from the H vector. (RevID(k) should not be changed when h ≠ RevID(k) in any event.)

We further modify these operations because we do not want to drop a variable $x_k$ selected in a Post-Ascent Phase when $x_k$ satisfies the recency threshold, giving it a high priority to be selected. Such an $x_k$ will be identified by S1 = True and hence, to avoid dropping such a high priority variable, we do not add $x_k$ to the After list when S1 = True. (A variation is to avoid adding $x_k$ when Condition1 = True. This reflects the situation where a variable with an S+ status is also given priority to avoid reversing its new assignment. The result is to execute fewer Add Steps and Drop Steps and does not correct for the myopic tendency as frequently.)

The Preliminary Initialization and the Post Iteration Update routines described in Section 6 contain the instructions to handle the required modifications introduced by the Myopic Correction Method. These refer to two Myopic Correction Inserts which are labeled MC1 and MC2. Based on the preceding observations, these two inserts are as follows.

**Insert MC1 for the Preliminary Initialization and the Post Iteration Update routines**
MovesAdded = 0; EndLink = BeginLink = 0; After(BeginLink) = EndLink; hLast = $h_o$.
H(h) = h, h = 1, …, $h_o$.

**Insert MC2 for the Post Iteration Update routine**
    *(Note: Updates for changing the value of $x_k$ and updating Eval(j) have been completed in the*
      *Post Iteration Update routine before executing the Add Step in this MC2 insert.*
      *Corresponding updates are additionally made in the Drop Step below when h = RevID(k)*
      *indicates that $x_k$ must change its value.)*
    *Execute **Add Step***
    If S1 = False then
        MovesAdded = MovesAdded + 1
        If hLast > 0 then
            h = H(hLast);  ID(h) = k; After(EndLink) = h; EndLink = h; RevID(k) = h;
            hLast := hLast − 1
        Endif
        If MovesAdded ≥ AddNum then
            *Execute **Drop Step***
            For Move = 1 to DropNum
                h = After(BeginLink); k = ID(h)
                If h = RevID(k) then
                      *(reverse the value of $x_k$)*
                      $x_k^\# := 1 - x_k^\#$
                      $x_o^\# := x_o^\# + Eval(k)$
                      EE(k) := EEbase − EE(k)
                      TabuIter(k) := Small + Iter

```
                        Update Eval(j) for all j = 1 to n
                Endif
                After(BeginLink) = After(h); hLast := hLast + 1; H(hLast) = h
            Endfor
            MovesAdded = 0
        Endif
    Endif
```

*Memory advantage of using After(h) instead of After(k)*

It is easy to see that if in general AddNum/DropNum ≥ 5, so that 1 variable is dropped only after 5 or more Add Steps, then more than $h_o + h_o/5$ Add Steps will be required to reach a point where hLast = 0 (the point at which the H list is "empty" and the After list contains $h_o$ elements). For subsequent Drop Steps to draw upon (and deplete) all these $h_o$ elements, an additional $5h_o$ Add Steps will be executed. (During these steps, the new $x_k$ variable will not be recorded each time hLast = 0.) Consequently, more than $6h_o$ moves will be required before some $x_k$ that should have been added is not found when accessing the After list in a Drop Step. Hence, even if an Ascent Phase or a Post-Ascent Phase consumes n moves, the value $h_o$ can be chosen as small as n/6 and the linked list will still allow all of these n moves to be accessed when performing Drop Steps. Normally, $h_o$ can be chosen somewhat smaller, for example n/12 if an Ascent or Post-Ascent Phase takes as many as n/2 moves, which is still likely excessive when n is large. If $h_o$ is chosen too small to handle all moves without a gap, it simply means that the Drop Steps will eventually skip over some of the $x_k$ variables that would otherwise have been added during an Add Step.

## 10. Concluding Observations and Future Steps

The departure from the classical approaches for responding to local optimality in the strategies of the AA algorithm open a variety of possibilities for exploration. Questions for experimentation concern whether one of the two versions of the Double-Pass Algorithm is superior to the other, and whether the value of the fraction F should be different for the two versions. The determination of preferred threshold parameters and the choice of values other than 2 for the exponential extrapolation parameter α, as discussed in the elaborations of Appendix 2, likewise invite investigation. Relevant questions include:

- What are the tradeoffs between r and Trigger of the recency and trigger thresholds? (Do the best values for each lie in some middle range and then deteriorate on either side of this range? Does the best value for Trigger (or for StatusCoun1), become smaller, perhaps as small as 1, as r becomes larger?
- Does an α value less than 2 become more effective as r or Trigger become larger?
- Are there advantages to joining path relinking (e.g, [9] and [10]) with the AA algorithm?
- What effect does complementing the values of the $x_j$ variables have in an intensification or diversification strategy (e.g., so that the best solution or a random solution is treated as the solution x = 0)?

- What contribution does the Myopic Correction Method make to the performance of the algorithm?
- Do answers to these questions depend on the state of the search, e.g., on how many iterations have elapsed or on how many Ascent and Post-Ascent phases have been performed?

Answers to these questions are also relevant to taking advantage of the flexibility of the AA algorithm to explore variants that are tailored for different classes of problems.

## Appendix 1: Variations and Extensions

The observations of this appendix and the next amplify considerations treated in the body of the paper.

### A1.1 Choosing parameter values for the recency, cutoff and trigger thresholds

Different values for r in the recency threshold $EE(j) \geq Threshold(r)$ can be suggested by tracking the value MaxEE in the Single-Pass and Double-Pass algorithms. A situation where MaxEE is consistently somewhat greater than Threshold(r) may be an indication that the value of r should be increased. However, it may be a mistake to choose r large by reference to MaxEE independent of when it is computed because there are likely to be variables $x_j$ with large $EE(j)$ values that are not desirable to be selected as $x_k$ and that fall under Condition 2 with $Eval(j) \leq 0$. Basing r on such values could then exclude the choice of desirable variables that satisfy the recency threshold with $Eval(j) > 0$. Consequently, if MaxEE is used as a guideline for selecting r, the value of MaxEE should be restricted to those cases where $Eval(j) > 0$.

The recency threshold might be adapted to differentiate the cases of using $EE1(j)$ and $EE0(j)$ by having a different r value for these two instances: for example, replacing (4.1) and (4.2) by

$$EE1(j) \geq Threshold(r1) \tag{4.1a}$$

and

$$EE0(j) \geq Threshold(r0). \tag{4.2a}$$

In many problems an optimal solution will have more variables equal 0 than 1, and this asymmetry may imply that it is preferable to choose $r0 > r1$ (or vice versa, if more variables equal 1 than 0).

For the inequalities defining the cutoff thresholds, there may be merit in choosing the fraction F larger (for example, closer to .5) when Condition1 = True, and perhaps larger still when S1 = True, because in these cases a somewhat smaller range of $x_j$ variables are candidates to be selected for $x_k$. For example, setting F = .7 when restricting attention to variables with Eval(j) > 0, and setting F = .5 when additionally restricting attention to variables satisfying the recency threshold, may roughly correspond to setting F = .9 when considering all variables without restriction.

Finally, the use of StatusCount1 and StatusCount2 to trigger an Ascent Phase is also subject to variation by modifying the trigger threshold inequality. For example, the S1 strategy may conceivably be more useful than the S2 strategy and applying it alone would evidently produce a simpler method. However, it seems plausible that some combination of the S1 and S2 strategies will work better than either one in isolation. Greater emphasis can nevertheless be placed the S1 strategy by a rule that launches an Assent Phase if either StatusCount1 + StatusCount2 ≥ Trigger or StatusCount1 ≥ Trigger1, where Trigger1 is slightly larger than Trigger/2.

These observations about variations and parameter values for the threshold strategies are related to, and can be influenced by, choices for α other than α = 2, as discussed in Appendix 2.

## A1.2 Alternative to the inductive formula and more general exponential extrapolation

Rather than use the inductive formula (5) from the beginning, starting from s = 1, it is also possible to accumulate the EE1(j) value for q = 1 to s, by starting for s = 1 with

$$EE1(j) = w(1)x_j$$

where $x_j$ is the value x(1,j) in the first local optimum x(1) and w(1) = 1. Then for each successive s = 1 to Q, if $EE1^p(j)$ represents the value of EE1(j) for the previous local optimum, then the current value of EE1(j) is

$$EE1(j) = w(s)x_j + EE1'(j) = 2^{s-1}x_j + EE1^p(j)$$

where $x_j$ is the value x(s,j) in the current local optimum x(s). Once s = Q, the formula (5) can then be used on all subsequent steps to update EE1(j).

This approach can be used with the more general exponential extrapolation formula (1)

$$w(q+1) = \alpha w(q) + \beta q + \gamma$$

This approach can grow Q to a selected value MaxQ, and then perform a diversification step such as the focal distance diversification strategy of Glover and Lu (2020) to start over again with Q = 1. By using more than one set of values for the parameters α, β and γ (or even just changing the value of the parameter α as in the strategies with β = γ = 0), these parameters can be applied for different MaxQ values, so that when one set is renewed by starting over another set will continue to apply to earlier local optima until its MaxQ value is reached. This "staggered" approach can then permit a parameter set that is renewed before another one to continue in operation when the second is renewed, and so on, so that there is always a connection to local optima from a relatively long period ago. In the case where $\beta = \gamma = 0$, which permits the inductive update, the use of real arithmetic allows the "effective" value of MaxQ to be quite large while keeping Q constant.

**A1.3 A Supplemental Approach Using Scaled Evaluations**

It may be useful to employ *scaled evaluations* as a supplemental method for identifying other forms of Rules 1 and 2 in Section 1.

As in Version 1 of the Double-Pass Algorithm, let MinEval, MaxEval and MeanEval refer to the min, max and means of the evaluations Eval(j), $j \in N$, on the current iteration. Suppose that target analysis or other experimentation discloses that the moves that yield correct new values $x_j = 1 - x_j^\#$ (corresponding to best known values $x_j^{best}$) occur for evaluations in the range LowEval to HighEval, where these values are chosen to maximize the number of correct new assignments that yield $1 - x_j^\# = x_j^{best}$, subject to limiting the number of incorrect new assignments where $1 - x_j^\# \neq x_j^{best}$. For example, if NumCorrect and NumIncorrect refer to the numbers of correct and incorrect assignments, then we may choose LowEval and HighEval to maximize NumCorrect – NumIncorrect or NumCorrect – 2·NumIncorrect, etc.

Then LowEval and HighEval are translated into associated scaled values, LowScaleV and HighScaleV that fall in the range from 0 to 100, expressed in relation to MinEval, MaxEval and MeanEval. Assume the scaled value ScaleV = 0 corresponds to MinEval, ScaleV = 100 corresponds to MaxEval, and ScaleV = 50 corresponds to MeanEval. Then we can identify ScaleV = LowScaleV and HighScaleV by the relation of Eval = LowEval and HighEval to MinEval, MaxEval and MeanEval as follows.

If Eval = MeanEval then ScaleV = 50 (7a)
If MinEval ≤ Eval < MeanEval then
    ScaleV = 50(Eval – MinEval)/(MeanEval – MinEval) (7b)
If MeanEval < Eval ≤ MaxEval then
    ScaleV = 50 + 50(Eval – MeanEval)/(MaxEval – MeanEval) (7c)

Let LowScaleV = ScaleV when Eval = LowEval and HighScaleV = ScaleV when Eval = HighEval, and let LowScaleV(Iter) and HighScaleV(Iter) refer to these values on a given iteration Iter. We collect these LowScaleV(Iter) and HighScaleV(Iter) values over a collection of iterations Iter ∈ IterSet, where for example IterSet refers to iterations where a succession of improving moves leads to a local optimum or where a succession of non-improving moves leads away from a local optimum. If the target analysis finding of [5] holds true about the difference between moves leading toward and away from local optima holds, then these two IterSets will likely contain somewhat different LowScaleV(Iter) and HighScaleV(Iter) values from each other, and LowScaleV(Iter) and HighScaleV(Iter) will likely be somewhat closer to 100 when leading toward a local optimum than when leading away from a local optimum.

Note that if we want to assure that IterSet refers to moves that lead unvaryingly toward a local optimum, we would additionally require that Eval > 0 in computing MinEval, MaxEval and MeanEval, hence restricting MinEval, MaxEval and MeanEval to refer to the min, max and mean of the evaluations Eval(j), $j \in N$ such that Eval(j) > 0. This could cause the scaled values LowScaleV(Iter) and HighScaleV(Iter) to lie closer to each other and change the relationship between these values and the values when IterSet refers to moving away from a local optimum. To make these two cases more nearly comparable, when moving away from a local optimum we would restrict MinEval, MaxEval and MeanEval to refer to the min, max and mean of the evaluations Eval(j), $j \in N$ such that Eval(j) ≤ 0.

Experimentation (such as target analysis) would then seek a representative pair LowScaleV and HighScaleV from the collection of LowScaleV(Iter) and HighScaleV(Iter) values for Iter $\in$ IterSet. For example, let MeanLowScaleV be the mean of the LowScaleV(Iter) values and let MeanHighScaleV be the mean of the HighScaleV(Iter) values over Iter $\in$ IterSet. Then LowScaleV might be chosen to equal MeanLowScaleV accompanied by selecting HighScaleV as the HighScaleV(Iter) value closest to MeanHighScaleV that would yield HighScaleV > LowScaleV. Similarly, HighScaleV might be chosen to equal MeanHighScaleV accompanied by selecting LowScaleV as the LowScaleV(Iter) value closest to MeanLowScaleV that would yield LowScaleV < HighScaleV.

Once such universal LowScaleV and HighScaleV values are selected, then the choice rules of the algorithm can be modified to select moves whose evaluations fall within these chosen ranges.

A different approach would seek to identify an improved type of aspiration criterion by addressing the question of how often tabu restrictions prevent good choices. Define *residual tabu tenure* to be the quantity TabuIter(ReverseMove) – CurrentIter. Then a criterion may be sought that is based on residual tabu tenure, the current $x_o$ value and the frequency that Eval(j) has been close to MaxEval (for instance, with ScaleV ≥ .95), and using experimentation involving target analysis to identify such a criterion whose choices are correlated with $1 - x_j^\# = x_j^{best}$.

Upon identifying the LowScaleV and HighScaleV values, the AA algorithm can be applied by translating LowScaleV and HighScaleV back into evaluations LowEval and HighEval for the current iteration. This may be done as follows. For the case where ScaleV is chosen to correspond to LowScaleV, set Eval = LowEval, and for the case where ScaleV is chosen to correspond to let HighScaleV, set Eval = HighEval. Then we have the following outcomes.

If ScaleV = 50 then Eval = MeanEval (8a)
If 0 ≤ ScaleV < 50 then
$$\text{Eval} = \text{MinEval} + (\text{MeanEval} - \text{MinEval})\text{ScaleV}/50 \quad (8b)$$

If 50 < ScaleV ≤ 100 then

$$\text{Eval} = \text{MeanEval} + (\text{MaxEval} - \text{MeanEval})(\text{ScaleV} - 50)/50 \qquad (8c)$$

(For example, if ScaleV = 75 then Eval = MeanEval + .5(MaxEval – MeanEval).)

### A.1.4 Efficient updates of evaluations.

As a supplementary observation to put some of the elements of the preceding sections in context, it may be noted that different problem settings afford different options for updating evaluations and this can influence an approach for updating evaluations efficiently. In some settings it is possible to update evaluations independently of each other, as in linear and mixed integer programming or network optimization where an update of a basis (or basis tree) allows different variables to change their values by a simple formula without the need to generate the values for other variables. This makes it possible to select a move at each iteration by examining only a chosen subset of variables, as in selecting a first improving move or a best improving move from a small sample. An advantage of such updates appears for problems with large numbers of variables where not all variables must be evaluated in order to make a choice on the current iteration.

By contrast, an efficient update in other settings may be provided by updating the evaluation for each variable on every iteration. This situation arises, for example, in updating Eval(j) for QUBO problems, where the new evaluation of each variable $x_j$ is calculated directly from the knowledge of Eval(j) on the preceding iteration. This type of approach has the advantage that the time required to identify a highest evaluation move is not dramatically greater than the time required to identify a first improving move.

In such a setting, where the evaluation Eval(j) of all variables must be updated after each iteration for efficiency, problems with large numbers of variables can usefully be handled by parallel processing that simultaneously updates different subsets of variables based on their previous evaluations. The strategies embodied in the AA algorithm introduce an important additional consideration by the Conditions 1 and 2, that cause evaluations to be treated differently. When parallel processing is used, if a variable satisfying Condition 1 is found within one of the processes, the fact that Condition 2 becomes irrelevant makes it advantageous to send a signal to the other processes so that all subsequent variables examined by these processes that satisfy Condition 2 can be ignored. The situation is compounded whenever a variable is discovered with an S1 status, enabling all future variables to be ignored unless they too have an S1 status. Again, the ability to signal other processes that an S1 status has been discovered can accelerate the algorithm's execution.

## Appendix 2. Implications of the recency threshold for using different α values

We begin by reviewing the meaning of the more general form of recency threshold EE1(j) ≥ Threshold(r) of (4.1) when α is not restricted to α = 2. Rewriting (3.3) with α replacing 2 gives

$$EE1(j) = \sum \alpha^{q-1} x_j^q : q = 1, \ldots, Q).$$

Expressing this as in the derivation of (5) yields

$$EE1(j) = \alpha^{Q-1} x_j^Q + \alpha^{Q-2} x_j^{Q-1} + \ldots + 1 x_j^1$$

Then the corresponding form the recency threshold of (4.1) becomes

$$\text{Threshold}(r) = \alpha^{Q-1} + \alpha^{Q-2} + \ldots + \alpha^{Q-r}.$$

We want to know the nature of the vector V(j) that satisfies EE1(j) ≥ Threshold(r), given by

$$V(j) = (x_j^Q, x_j^{Q-1}, \ldots, x_j^1)$$

Proceeding by example, suppose we choose Q = 7 and r = 3, giving

$$\text{Threshold}(3) = \alpha^6 + \alpha^5 + \alpha^4.$$

We examine the α values 1.7 and 1.5 to compare with α = 2. The sequence $(\alpha^q) = \alpha^{Q-1}, \alpha^{Q-2}, \ldots, \alpha^1, \alpha^0$ for Q = 7, and for α = 2, 1.7 and 1.5 becomes as follows (rounding off to 4 decimal places):

**Weight Table**

| q = | 6 | 5 | 4 | 3 | 2 | 1 | 0 |
|---|---|---|---|---|---|---|---|
| $2^q =$ | 64 | 32 | 16 | 8 | 4 | 2 | 1 |
| $1.7^q =$ | 24.1376 | 14.1986 | 8.3521 | 4.913 | 2.89 | 1.7 | 1 |
| $1.5^q =$ | 11.3906 | 7.5937 | 5.0625 | 3.375 | 2.25 | 1.5 | 1 |

The preceding Weight Table discloses that for α = 2, 1.7 and 1.5 we obtain Threshold(3) = 112, 46.6883 and 24.0468, respectively.

We call the vector V(j) *acceptable* if it satisfies the recency threshold EE1(j) ≥ Threshold(3) and let # denote the option of either # = 0 or 1. First, note that the only acceptable V(j) vectors for α = 2 have the form

$$V(j) = (1\ 1\ 1\ \#\ \#\ \#\ \#).$$

This means that $x_j = 1$ in each of the 3 most recent local optima $x(Q)$, $x(Q-1)$ and $x(Q-2)$ (i.e., $x_j^Q = 1$, $x_j^{Q-1} = 1$ and $x_j^{Q-2} = 1$), while $x_j = 0$ and $x_j = 1$ are both possible in earlier local optima $x(Q-3)$ to $x(1)$.

Requiring $EE1(j) \geq Threshold(3)$ therefore compels $x_j = 1$ in the 3 most recent local optima when $\alpha = 2$. When $\alpha < 2$, other $V(j)$ vectors in addition to $V(j) = (1\ \ 1\ \ 1\ \ \#\ \ \#\ \ \#\ \ \#)$ can satisfy the recency threshold. Consequently, in some cases $x_j = 1$ may not be required for each of the 3 most recent local optima.

**Acceptable V(j) vectors**

By considering the subsets of q values in the Weight Table that permit $EE1(j) \geq Threshold(3)$ to be satisfied in each of the cases $\alpha = 2, 1.7$ and $1.5$, we obtain the following outcomes.

For $\alpha = 2$: $(1\ \ 1\ \ 1\ \ \#\ \ \#\ \ \#\ \ \#)$

For $\alpha = 1.7$: $(1\ \ 1\ \ 1\ \ \#\ \ \#\ \ \#\ \ \#)$, $(1\ \ 1\ \ 0\ \ 1\ \ 1\ \ 1\ \ \#)$, $(1\ \ 1\ \ 0\ \ 1\ \ 1\ \ 0\ \ 1)$
    (3 more options than for $\alpha = 2$, accounting for $\# = 0$ or $1$)

For $\alpha = 1.5$: $(1\ \ 1\ \ 1\ \ \#\ \ \#\ \ \#\ \ \#)$, $(1\ \ 1\ \ 0\ \ 1\ \ 1\ \ \#\ \ \#)$, $(1\ \ 1\ \ 0\ \ 1\ \ 0\ \ 1\ \ 1)$,
    $(1\ \ 0\ \ 1\ \ 1\ \ 1\ \ 1\ \ 1)$  (6 more options than for $\alpha = 2$, accounting for $\# = 0$ or $1$)

To further see the relevance of these differences, recall that Strategy S1 uses the recency threshold $EE1(j) \geq Threshold(r)$ when the $x_j = 1$ in the most recent local optimum ($x_j^Q = 1$ in $x(Q)$), and we want to decide whether to change $x_j$ to give $x_j = 0$ (under conditions where this change is evaluated to improve the current solution). As previously emphasized, when $\alpha = 2$, changing $x_j$ to give $x_j = 0$ causes $x_j$ to take a different value than in the 3 most recent local optima, and hence we will not duplicate any of these local optima as long as $x_j$ retains its new value of 0. (Analogously, to decide whether to change $x_j = 0$ to $x_j = 1$, we use the inequality (4.2) $EE0(j) \geq Threshold(r)$ that replaces $EE1(j)$ by $EE0(j)$.)

When $\alpha = 1.7$ above, the solutions $(1\ \ 1\ \ 0\ \ 1\ \ 1\ \ 1\ \ \#)$ and $(1\ \ 1\ \ 0\ \ 1\ \ 1\ \ 0\ \ 1)$ show that changing $x_j = 1$ to $x_j = 0$ would cause the new solution to have a different value than in the two most recent local optima (where $x_j^Q = x_j^{Q-1} = 1$), but there are three cases where changing $x_j = 1$ to $x_j = 0$ would yield the same $x_j$ value as in the third most recent local optimum (where $x_j^{Q-2} = 0$ in these solutions). Consequently, there would be a possibility that changing $x_j = 1$ to $x_j = 0$ would permit the third most recent local optimum to be revisited. This possibility might not be large, considering that most of the local optima avoided by $\alpha = 1.7$ are represented by the solutions $(1\ \ 1\ \ 1\ \ \#\ \ \#\ \ \#\ \ \#)$. The risk of revisiting the $r^{th}$ most recent solution would also clearly have a smaller impact if r is somewhat greater than 3. The risk would further be diminished if other variables $x_j$ likewise satisfied the recency threshold, since each of these instances would mostly avoid the solutions represented by $(1\ \ 1\ \ 1\ \ \#\ \ \#\ \ \#\ \ \#)$.

The case for α = 1.5 shows this smaller α value poses additional risks beyond α = 1.7 of revisiting solutions other than (1  1  1  #  #  #  #). One of these risks duplicating the second most recent local optimum. (Since this solution is the one indexed x(r – 1), the significance of this risk is not very great as r becomes larger.)

In all of these cases, the risk might be additionally reduced as the number of moves away from the most recent local optimum increases, since this produces a chance that the ascent to a new local optimum would be launched from a point farther away from previous local optima. However, greater assurance would be provided by the trigger threshold that postpones the Ascent Phase until an increased number of different $x_k$ variables are identified by Strategies S1 and S2 whose V(k) vectors satisfy EE1(k) ≥ Threshold(r).

As in the case of α = 2, it is not necessary to record these V(k) vectors, since the simple update of EE1(j) for all j can be used with the general form of (5) where α replaces 2; i.e.,

$$EE1(j) = \alpha^{Q-1} x_j^Q + EE1(j){:}p/\alpha$$

By these observations it is clear that there may be merit in exploring the use of α values other than α = 2 when exponential extrapolation is embedded in an adaptive memory strategy. For example, the preceding examples show that smaller α values can avoid revisiting some local optima beyond the first r, and this might be additionally exploited by choosing larger r values for smaller α values. The chief appeal of using an α value less than 2 is that it allows greater latitude in the choice of variables that qualify for launching a new Ascent Phase by Strategy S1 or S2.

## Appendix 3: Complementarity relationships and the complementary recency threshold

It is useful to define the complement $Eval^c(j)$ of Eval(j), where Eval(j) is the evaluation of $x_j = x_j^\#$ and $Eval^c(j)$ is the evaluation of the complementary assignment $x_j = 1 - x_j^\#$ by

$$Eval^c(j) = -\ Eval(j)$$

This corresponds to the observation that Eval(j) changes to – Eval(j) when $x_j$ changes from $x_j = x_j^\#$ to $x_j = 1 - x_j^\#$. (The effect on $x_o$ that yields $x_o^\# := x_o^\# + Eval(j)$ when $x_j$ changes from $x_j^\#$ to $1 - x_j^\#$ must be reversed to give back the original value of $x_o$ when $x_j$ is again changed from $1 - x_j^\#$ to $x_j^\#$, hence the latter change must give $x_o^\# := x_o^\# + Eval(j) - Eval(j) = x_o^\#$.)

Then, just as EE(j) refers to the assignment $x_j = x_j^\#$ (with an exception in defining S2 status) we define its complement $EE^c(j)$ to refer to the assignment $x_j = 1 - x_j^\#$ by

$$EE^c(j) = EEbase - EE(j)$$

This also gives $EE(j) = EEbase - EE^c(j)$ and results from the fact that $EE0(j) = EEbase - EE1(j)$ and conversely.

Finally, we note that Threshold(r) is independent of the value assigned to $x_j$ and define its complement by

$$Threshold^c(r) = EEbase - Threshold(r)$$

which similarly yields $Threshold(r) = EEbase - Threshold^c(r)$.

From these definitions it may be verified that the recency threshold $EE(j) \geq Threshold(r)$ of (4) gives rise to the *complementary recency threshold* (in the opposite direction)

$$EE^c(j) \leq Threshold^c(r) \tag{4$^c$}$$

The significance of (4$^c$) is that whenever the recency threshold $EE(j) \geq Threshold(r)$ of (4) holds and $x_j$ is chosen to change its value from $x_j^\#$ to $1 - x_j^\#$, after the assignment, for the new value of $x_j$, we will have

$$EE(j) \leq Threshold^c(r) = EEbase - Threshold(r) \tag{4-alt}$$

From the definitions $EEbase = \sum(w(q): q = 1, \ldots, Q)$ $(= 2^{Q-1} + 2^{Q-2} + \ldots + 2^0)$ and $Threshold(r) = 2^{Q-1} + 2^{Q-2} + \ldots + 2^{Q-r}$, the quantity $Threshold^c(r)$ can also be written

$$Threshold^c(r) = 2^{Q-r-1} + \ldots + 2^0$$

which is evidently much smaller than Threshold(r) (since $2^{Q-r} > Threshold^c(r)$ by the relationship $2^{Q-r} = (2^{Q-r-1} + \ldots + 2^0) + 1$).

Hence when the recency threshold is satisfied for $x_j = x_j^\#$, (4-alt) implies the threshold cannot be satisfied after changing $x_j$ to $1 - x_j^\#$. The converse is also true, if $EE(j) \leq Threshold^c(r)$ is satisfied for $x_j = x_j^\#$, then the recency threshold will be satisfied when $x_j$ is changed to equal $1 - x_j^\#$.

## Appendix 4: Tradeoff Relationships

A refers to a current evaluation and B refers to a previous evaluation, such as the best before now.

$A_1$ and $B_1$ refer to the first type of evaluation and $A_2$ and $B_2$ refer to the second type of

evaluation. We assume the second type of evaluation, A2 and B2, is always nonnegative (as in the case of EE(j)), but the first type, $A_1$ and $B_1$, can sometimes be negative (as in the case of Eval(j)).

The current evaluation will dominate the previous evaluation if $A \geq B$; i.e.,

$$A_1 \geq B_1 \text{ and } A_2 \geq B_2.$$

Assume dominance does not occur. Then we have two possibilities.

**Case 1**. $A_1 > B_1$ and $A_2 < B_2$

**Case 2**. $A_1 < B_1$ and $A_2 > B_2$

Consider these two cases in the context of Condition 1 and Condition 2, which we write as follows:

**Condition 1**. $A_1, B_1 \geq 0$. $A_2, B_2 \geq 0$

**Condition 2**. $A_1, B_1 \leq 0$. $A_2, B_2 \geq 0$.

These conditions correspond to Conditions 1 and 2 of Section 6 where $A_1$ and $B_1$ refer to Eval(j) and $A_2$ and $B_2$ refer to EE(j). However, the conditions here are less stringent than those of section 6, since they do not include reference to tabu restrictions or the recency threshold or the S1 status of variables. In addition, Condition 1 of Section 6 would imply $A_1, B_1 > 0$ rather than $A_1, B_1 \geq 0$. We note, however, that we can translate every case for Condition 2 into Condition 1 by identifying a lower bound LB for all instances $A_1$ and $B_1$ such that $A_1, B_1 \geq LB$, and redefining

$$A_1 := A_1 - LB; \quad B_1 := B_1 - LB.$$

Without identifying LB, we consider Conditions 1 and 2 separately. For each combination of conditions and cases, we identify the max and min values of the A and B components.

**Condition 1 & Case 1**.

The combination of Condition 1 and Case 1 yields $A_1 > B_1 \geq 0$, hence $A_1 > 0$, and we seek a nonnegative multiple x so that $A_1x$ dominates $B_1$, as given by
$$A_1x \geq B_1$$

We also have $B_2 > A_2 \geq 0$, hence $B_2 > 0$, and we seek a nonnegative multiple x so that $A_2$ dominates $B_2x$, as given by

$$A_2 \geq B_2 x.$$

An x that yields dominance in both situations gives $A_2/B_2 \geq x \geq B_1/A_1$ or equivalently
$$A_1 A_2 / A_1 B_2 \geq x \geq B_1 B_2 / A_1 B_2$$

Hence dominance and strict dominance are respectively achieved by
$$A_1 A_2 \geq B_1 B_2 \text{ and } A_1 A_2 > B_1 B_2.$$

In terms of Eval(j) and EE(j) this corresponds to $\text{Eval}(j) \cdot \text{EE}(j) > \text{Eval}_p(j) \cdot \text{EE}_p(j)$, where the "p" subscript represents "previous".

**Condition 1 & Case 2**

Corresponding analysis gives
$$A_1 \geq B_1 x \text{ and } A_2 x \geq B_2$$

to yield
$$A_1 A_2 / A_2 B_1 \geq x \geq B_1 B_2 / A_2 B_1$$

and while the denominator is different, the conclusions for dominance and strict dominance are the same as in Condition 1 & Case 1, i.e.,
$$A_1 A_2 \geq B_1 B_2 \text{ and } A_1 A_2 > B_1 B_2.$$

**Condition 2 & Case 1**

We now have $0 \geq A_1 > B_1$, hence $B_1 < 0$, and we seek a nonnegative multiple x so that $A_1$ dominates $B_1 x$, hence
$$A_1 \geq B_1 x \text{ or } -B_1 x \geq -A_1.$$

Likewise, we have $B_2 > A_2 \geq 0$, hence $B_2 > 0$, and we seek a nonnegative multiple x so that $A_2$ dominates $B_2 x$, hence
$$A_2 \geq B_2 x.$$

Since $-B_1 > 0$, the two inequalities become $A_2/B_2 \geq x \geq -A_1/-B_1$
or equivalently
$$-A_2 B_1 / -B_1 B_2 \geq x \geq -A_1 B_2 / -B_1 B_2$$

with positive denominators. Hence dominance and strict dominance are achieved by
$$-A_2 B_1 \geq -A_1 B_2 \ (A_1 B_2 \geq A_2 B_1) \text{ and } -A_2 B_1 > -A_1 B_2 \ (A_1 B_2 > A_2 B_1)$$

In terms of Eval(j) and EE(j) this corresponds to $\text{Eval}(j) \cdot \text{EE}_p(j) > \text{Eval}_p(j) \cdot \text{EE}(j)$, where the "p" subscript again represents "previous".

**Condition 2 & Case 2**

Following the line of argument as in Condition 2 & Case 1, we conclude

$$-A_2B_1/A_1A_2 \geq x \geq -A_1B_2/A_1A_2$$

which yields the same dominance conclusions as in Condition 2 & Case 1.

We remark that the conclusions in all these cases can also be reached by a more involved derivation using a different definition of dominance, where A dominates B if

$$(\text{Max}(A_1, B_1) - \text{Min}(A_1,B_1)/(|A_1| + |B_1|) \geq (\text{Max}(A_2, B_2) - \text{Min}(A_2,B_2)/(|A_2| + |B_2|).$$

## Appendix 5. Better Aspiration Criteria

*(The following observations are drawn from the unpublished note "Decision Rule Enhancement with Metaheuristic Analysis". They are included here to identify a different situation in which it can be appropriate to remove tabu restrictions and proceed to a local optimum.)*

The customary aspiration criterion used with tabu search in binary optimization allows a tabu variable $x_j$, currently assigned the value $x_j = x_j^\#$, to be selected to change its value in spite of being tabu, provided the new assignment $x_j = 1 - x_j^\#$ will cause the objective function value $x_o$ to be better than the best value $x_o^*$ obtained so far. This criterion has an evident shortcoming in a situation where the tabu status of moves prevents choices of a sequence of assignments that would result in yielding $x_o = x_o^\# > x_o^*$, as can happen when one or more component moves of the sequence is tabu. This shortcoming can also prevent a solution x from being added to a reference set R because it is not better than the lowest ranking solution x' in R, but could become better than x' if the tabu restrictions were lifted and moves were allowed that would cause x to be improved beyond x'.

Consequently, it can be useful to employ a different aspiration criterion that can generate a *tabu free solution* which is obtained as a local optimum after removing all tabu restrictions. During an initial phase of search, tabu restrictions are normally not imposed, and hence the first local optimum reached is a tabu free solution. At this point, the last assignment $x_j = x_j^\#$ is typically identified to make the reverse assignment $x_j = 1 - x_j^\#$ tabu for a specified number of iterations denoted by TabuTenure, which can vary randomly about a value BaseTenure.

The question arises: when can tabu restrictions be removed to permit a tabu free solution to be obtained? Such a removal should not be done immediately, because then the search can fall back into the same local optimum previously found. Instead, an option is to use a parameter

TabuRange (analogous to the parameter for TabuTenure), where the choice of TabuTenure and TabuRange should be coordinated so that TabuRange > TabuTenure.

Let TabuFreeIter record the iteration when the last tabu free solution was generated. Then the traditional aspiration criterion of $x_o^\# > x_o^*$ can be used until the number of iterations reaches TabuFreeIter + TabuRange. From this point on, the aspiration criterion becomes $x_o^\# > x_o^* - x_o\text{Tolerance}$, where $x_o\text{Tolerance}$ is a positive value to permit tabu restrictions to be temporarily disregarded when $x_o^\# > x_o^* - x_o\text{Tolerance}$, and hence to proceed to a new tabu free solution (which may or may not be an improved best solution).

However, when tabu restrictions are temporarily disregarded upon satisfying $x_o^\# > x_o^* - x_o\text{Tolerance}$, these restrictions are not permanently forgotten, but reinstated when the new tabu free solution is obtained (accounting for the fact that the residual tabu tenures given by TabuIter(j) – CurrentIter will be reduced since new iterations have elapsed). In addition, the new moves used to reach the tabu free local optimum are also used to generate tabu restrictions in the normal way, and once the new local optimum is reached, all tabu restrictions are reinstated. (The gradual extinction of tabu tenures is handled in the usual fashion by setting TabuIter(j) = TabuTenure + CurrentIter, and then $x_j$ is tabu until CurrentIter grows to satisfy CurrentIter > TabuIter(j).)